\RequirePackage{snapshot}
\documentclass[10pt,twocolumn,letterpaper]{article}

\usepackage{titling}
\usepackage{cvpr}
\usepackage{times}
\usepackage{epsfig}
\usepackage{graphicx}
\usepackage{amsmath}
\usepackage{amssymb}

\usepackage{xifthen}
\usepackage{xcolor}
\usepackage{cite}

\makeatletter
\@namedef{ver@everyshi.sty}{}
\makeatother
\usepackage{tikz}

\newcommand{\rpm}{\sbox0{$1$}\sbox2{$\scriptstyle\pm$}
  \raise\dimexpr(\ht0-\ht2)/2\relax\box2 }

\newcommand{\PAR}[1]{\vskip4pt \noindent{\bf #1~}}

\usepackage[pagebackref=true,breaklinks=true,letterpaper=true,colorlinks,bookmarks=false]{hyperref}

\cvprfinalcopy %

\def\cvprPaperID{***} %

\ifcvprfinal\fi
\begin{document}

\title{Why Having 10,000 Parameters in Your Camera Model is Better Than Twelve}

\author{Thomas Sch\"{o}ps$^1$
\and
Viktor Larsson$^1$
\and
Marc Pollefeys$^{1,2}$
\and
Torsten Sattler$^3$
\and
$^1$Department of Computer Science, ETH Z\"{u}rich \quad  $^3$Chalmers University of Technology\\ 
$^2$Microsoft Mixed Reality \& AI Zurich Lab
}

\maketitle

\begin{abstract}
Camera calibration is an essential first step in setting up 3D Computer Vision systems. Commonly used parametric camera models are limited to a few degrees of freedom and thus often do not optimally fit to complex real lens distortion. In contrast, generic camera models allow for very accurate calibration due to their flexibility. Despite this, they have seen little use in practice. In this paper, we argue that this should change. We propose a calibration pipeline for generic models that is fully automated, easy to use, and can act as a drop-in replacement for parametric calibration, with a focus on accuracy. We compare our results to parametric calibrations. Considering stereo depth estimation and camera pose estimation as examples, we show that the calibration error acts as a bias on the results. We thus argue that in contrast to current common practice, generic models should be preferred over parametric ones whenever possible. To facilitate this, we released our calibration pipeline at {\small{\url{https://github.com/puzzlepaint/camera_calibration}}}, making both easy-to-use and accurate camera calibration available to everyone.
\end{abstract}

\vspace{-1em}

\section{Introduction}
\begin{figure}
\begin{center}
\includegraphics[trim={12px 12px 12px 12px},clip,width=0.33\linewidth]{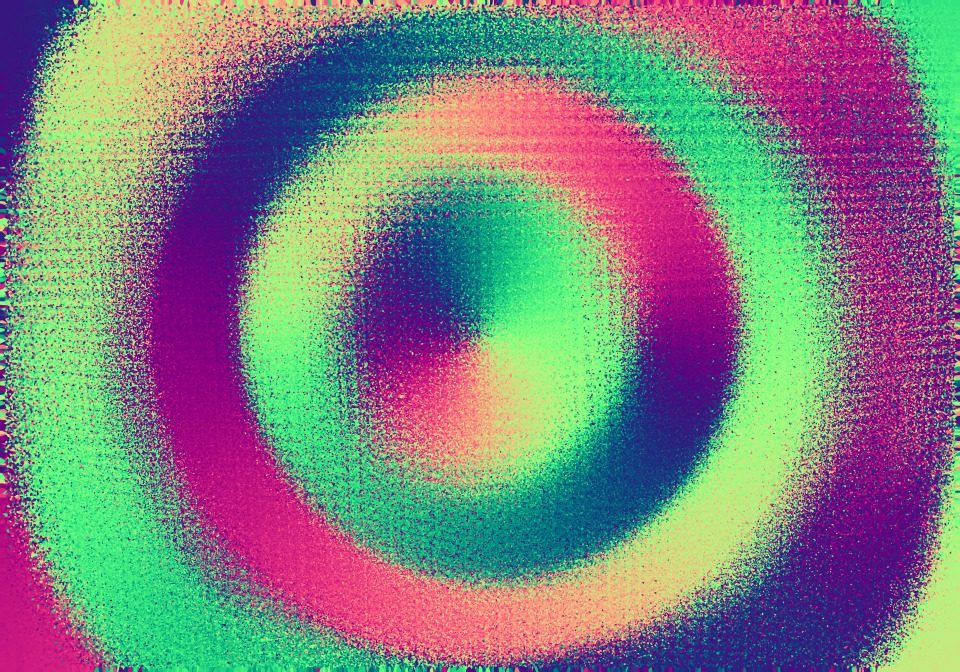}%
\includegraphics[trim={12px 12px 12px 12px},clip,width=0.33\linewidth]{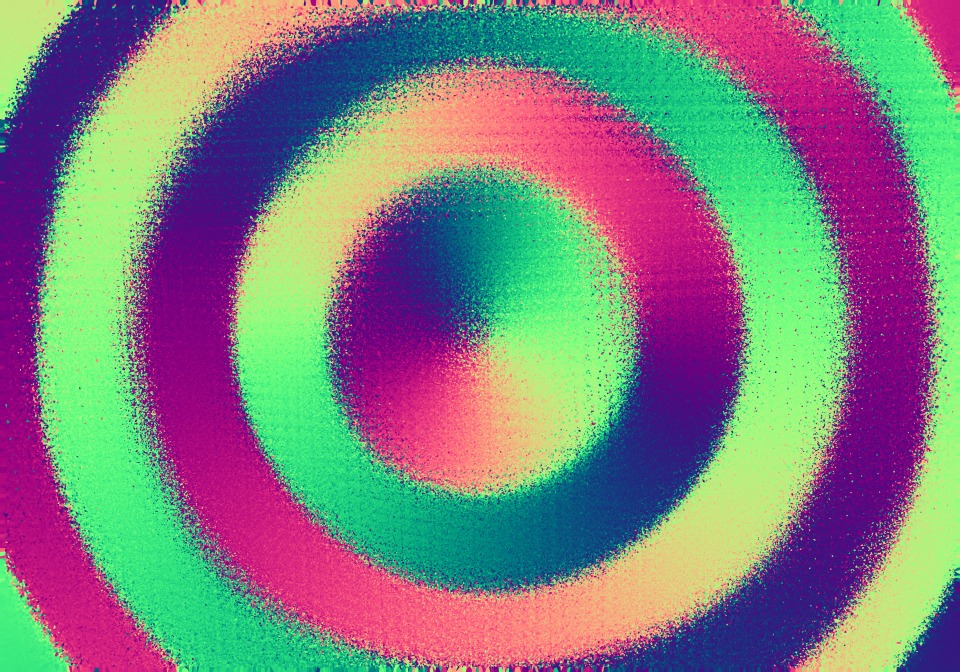}%
\includegraphics[trim={12px 12px 12px 12px},clip,width=0.33\linewidth]{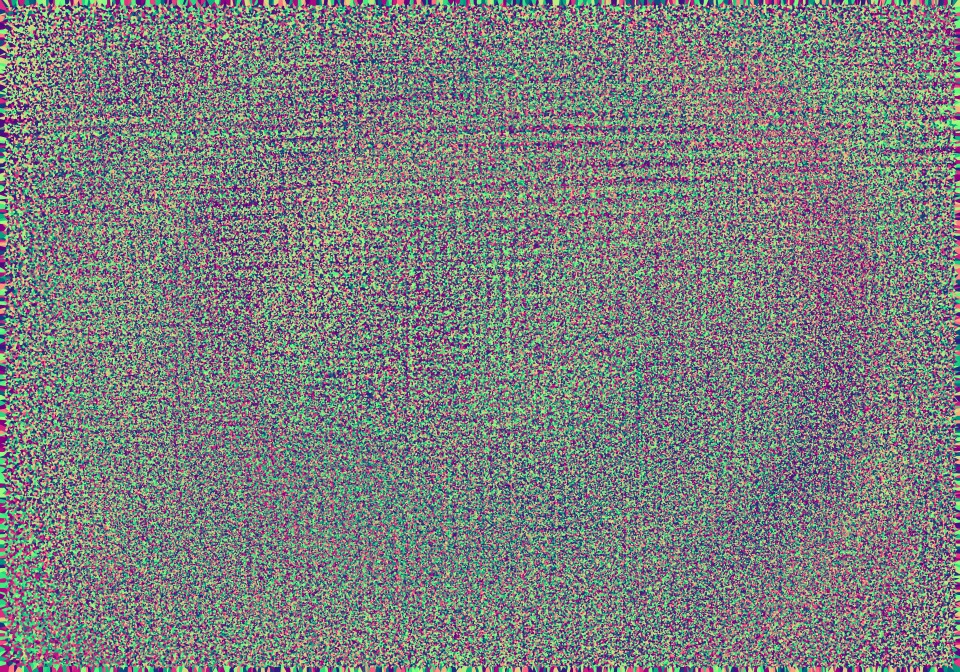}%
\end{center}
\vspace{-1.0em}
\caption{
Residual distortion patterns of fitting two parametric camera models (left, center) and a generic model (right) to a mobile phone camera.
While the generic model shows mostly random noise, parametric models show strong systematic modeling errors.
\vspace{-1.9em}
}
\label{fig:teaser}
\end{figure}

Geometric camera calibration is the process of determining where the light recorded by each pixel of a camera comes from.
It is an essential prerequisite for 3D Computer Vision systems. 
Common parametric camera models allow for only a few degrees of freedom and are thus unlikely to optimally fit to complex real-world lens distortion (\cf Fig.~\ref{fig:teaser}).
This can for example be aggravated by placing cameras behind windshields for autonomous driving \cite{beck2018generalized}.
However, accurate calibration is very important, since calibration errors affect all further computations.
Even though the noise introduced by, for example, feature extraction in the final application is likely much larger than the error in the camera calibration, the latter can still be highly relevant since it may act as a \emph{bias} that cannot be averaged out.

\begin{figure}
\begin{center}
\includegraphics[trim={0 0 0 0},clip,width=0.99\linewidth]{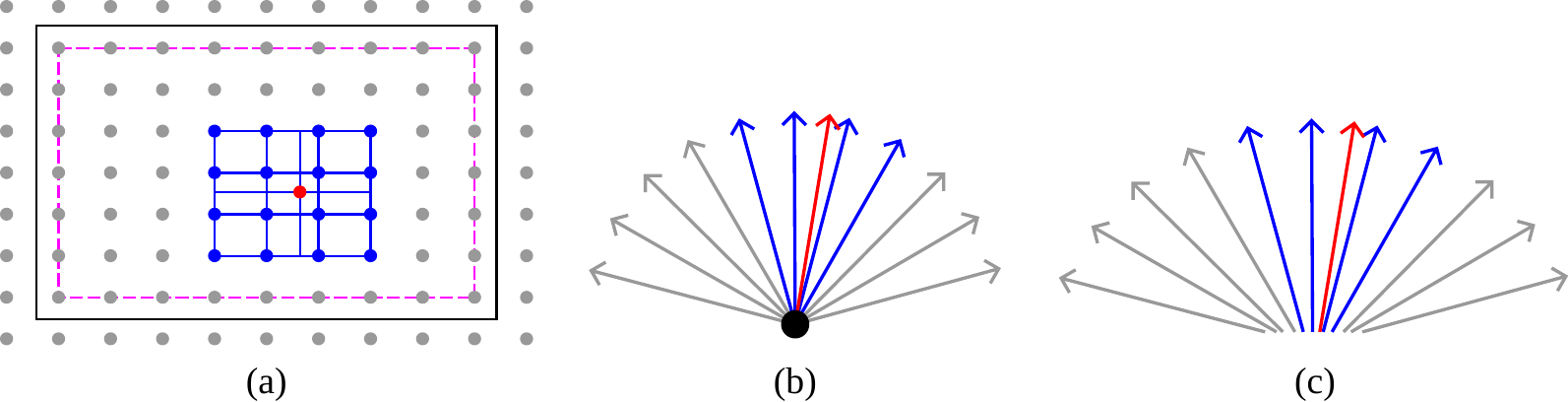}%
\end{center}
\vspace{-1.0em}
\caption{
2D sketch of the two generic camera models considered in this paper.
(a) In image space (black rectangle), a grid of control points is defined that is aligned to the calibrated area (dashed pink rectangle) and extends beyond it by one cell. 
A point (red) is un-projected by B-Spline surface interpolation of the values stored for its surrounding 4x4 points (blue). 
Interpolation happens among directions (gray and blue arrows) starting from a projection center (black dot) for the central model (b), and among arbitrary lines (gray and blue arrows) %
for the non-central model (c).
\vspace{-1.5em}
}
\label{fig:model_sketch}
\end{figure}

Generic %
camera models \cite{grossberg2001general} %
relate pixels and their 3D observation lines resp.~rays outside of the camera optics in a purely mathematical way, without offering a physical interpretation of the camera geometry.
They densely associate pixels with observation lines or rays; in the extreme case, a separate line is stored for each pixel in the camera image. %
Due to their many degrees of freedom, they may fit all kinds of cameras, allowing to obtain accurate, bias-free calibrations.
Fig.~\ref{fig:model_sketch} shows the models considered in this paper.

Previous generic calibration approaches (\cf Sec.~\ref{sec:related_work}) have seen limited use in practice.
On the one hand, this might be since there is no readily usable implementation for any existing approach.
On the other hand, the community at large does not seem to be aware of the practical advantages of generic calibration approaches over parametric models.

Our contributions are thus:
1) We propose improvements to camera calibration, in particular to the calibration pattern and feature extraction, to increase accuracy.
2) We show the benefits of accurate generic calibration over parametric models, in particular on the examples of stereo depth and camera pose estimation.
3) %
We publish our easy-to-use calibration pipeline and generic camera models as open source.

\section{Related Work}
\label{sec:related_work}
In this section, we %
present related work on calibration with generic camera models. %
We do not review calibration initialization, %
since we adopt \cite{ramalingam2016unifying} which works well for this. %

\PAR{Pattern design and feature detection.}
Traditionally, checkerboard \cite{camera_calibration_toolbox_matlab,opencv_library} and dot \cite{heikkila2000geometric} patterns have been used for camera calibration.
Feature detection in dot patterns is however susceptible to perspective and lens distortion \cite{mallon2007pattern}.
Recent research includes more robust detectors for checkerboard patterns \cite{placht2014rochade,duda2018accurate}, the use of ridge lines for higher robustness against defocus \cite{ding2017robust}, and calibration with low-rank textures \cite{zhang2011camera}.
Ha \etal \cite{ha2017deltille} propose the use of triangular patterns, which provide more gradient information for corner refinement than checkerboard patterns.
Our proposed calibration pattern similarly increases the available gradients, while however allowing to vary the black/white segment count, enabling us to design better features than \cite{ha2017deltille}.

\PAR{Non-central generic models.} 
Grossberg and Nayar~\cite{grossberg2001general} first introduced a generic camera model that associates each pixel with a 3D observation line, defined by a line direction and a point on the line. %
This allows to model central cameras, \ie, cameras with a single unique center of projection, as well as non-central cameras. 
\cite{ramalingam2016unifying,sturm2004generic} proposed a geometric calibration approach for the generic model from~\cite{grossberg2001general} that does not require known relative poses between images. 
\cite{ramalingam2016unifying} focus on initialization rather than a full calibration pipeline.
Our approach extends \cite{ramalingam2016unifying} with an improved calibration pattern / detector and adds full bundle adjustment. 

\PAR{Central generic models.}
Non-central cameras may complicate applications, \eg, %
undistortion to a pinhole image is not possible without knowing the pixel depths~\cite{Swaninathan2003CVPR}.
Thus, models which constrain the calibration to be central were also proposed \cite{dunne2010efficient,bergamasco2017parameter,beck2018generalized,Nister2005ICCV}. 
For a central camera, all observation lines intersect in the center of projection, simplifying observation lines to observation rays / directions. 


\PAR{Per-pixel models vs.~interpolation.}
Using a observation line / ray for each pixel~ \cite{grossberg2001general,dunne2010efficient} provides maximum flexibility. 
However, this introduces an extreme number of parameters, making calibration harder. 
In particular, classical sparse calibration patterns do not provide enough measurements. 
Works using these models thus obtain dense matches using displays that can encode their pixel positions \cite{grossberg2001general,dunne2010efficient,bergamasco2013can,bergamasco2017parameter}, or interpolate between sparse features~\cite{ramalingam2016unifying}. 

Since using printed patterns can sometimes be more practical than displays, and interpolating features causes inaccuracy \cite{dunne2010efficient}, models with lower calibration data requirements have been proposed.
These interpolate between sparsely stored observation lines resp.~rays.
\Eg, \cite{miraldo2012calibration} propose to interpolate arbitrarily placed control points with radial basis functions. 
Other works use regular grids for more convenient interpolation.
\cite{beck2018generalized} map from pixels to observation directions with a B-Spline surface.
\cite{rosebrock2012generic,rosebrock2012complete} use two spline surfaces to similarly also define a non-central model. 
In this work, we follow these approaches.

The above works are the most similar ones to ours regarding the camera models and calibration.
Apart from our evaluation in real-world application contexts, we aim to achieve even more accurate results. 
Thus, our calibration process differs as follows:
1) We specifically design our calibration pattern and feature detection for accuracy (\cf Sec.~\ref{sec:data_collection}, \ref{sec:feature_extraction}).
2) \cite{beck2018generalized,rosebrock2012generic} %
approximate the reprojection error in bundle adjustment. %
We avoid this approximation since, given Gaussian noise on the features, this will lead to better solutions. 
3) \cite{beck2018generalized,rosebrock2012generic} assume planar calibration patterns which will be problematic for imperfect patterns.
We optimize for the pattern geometries in bundle adjustment, accounting for real-world imperfections~\cite{Strobl2011ICCVW}. 
4) We use denser control point grids than \cite{beck2018generalized,rosebrock2012generic}, allowing us to observe and model interesting fine details (\cf Fig.~\ref{fig:model_eval_bias_free}).


\PAR{Photogrammetry.}
The rational polynomial coefficient (RPC) model \cite{hu2004understanding} maps 3D points to pixels via ratios of polynomials of their 3D coordinates.
With 80 parameters, it is commonly used for generic camera modeling in aerial photogrammetry. %
In contrast to the above models, its parameters globally affect the calibration, making it harder to use more parameters. 
Further, this model works best only if all observed 3D points are in a known bounded region.

\PAR{Evaluation and comparison.}
Dunne \etal \cite{dunne2007comparison} compare an early variant \cite{sturm2003generic} of Ramalingam and Sturm's series of works \cite{sturm2003generic,%
sturm2004generic,%
ramalingam2005theory,%
ramalingam2005towards,%
ramalingam2008minimal,%
ramalingam2016unifying} with classical parametric calibration \cite{sturm1999plane,zhang2000flexible}.
They conclude that the generic approach works better for high-distortion cameras, but worse for low- to medium-distortion cameras.
In contrast, Bergamasco \etal \cite{bergamasco2013can} conclude for their approach that even quasi-pinhole cameras benefit from non-central generic models.
Our results also show that generic models generally perform better than typical parametric ones, and we in addition evaluate how this leads to practical advantages in applications.

\section{Accurate Generic Camera Calibration}
The first step in our pipeline
is to record many photos or a video of one or multiple calibration patterns to obtain enough data for dense calibration (Sec.~\ref{sec:data_collection}).
We propose a %
pattern that enables very accurate feature detection.
The next step is to detect the features in the images (Sec.~\ref{sec:feature_extraction}).
After deciding for the central or non-central camera model (Sec.~\ref{sec:camera_model}), the camera is calibrated:
First, a dense per-pixel initialization is obtained using \cite{ramalingam2016unifying}.
Then, the final model is fitted to this and refined with bundle adjustment %
(Sec.~\ref{sec:calibration}).

All components build relatively closely on previous work, as indicated below; %
our contributions are the focus on accuracy in the whole pipeline, and using it to show the limitations of parametric models in detailed experiments. 
Note that our approach assumes that observation rays / lines vary smoothly among neighbor pixels, without discontinuities.

\subsection{Calibration Pattern \& Data Collection}
\label{sec:data_collection}
\vspace{-2pt}
For data collection, we record images of a known calibration pattern.
This allows for virtually outlier-free and accurate localization of feature points on the pattern.
Thus, compared to using natural features, less data is required to average out errors.
Furthermore, the known (near-planar) geometry of the pattern is helpful for initialization.

\begin{figure}
\begin{center}
\includegraphics[trim={1em 1em 1em 1em},clip,height=0.35\linewidth]{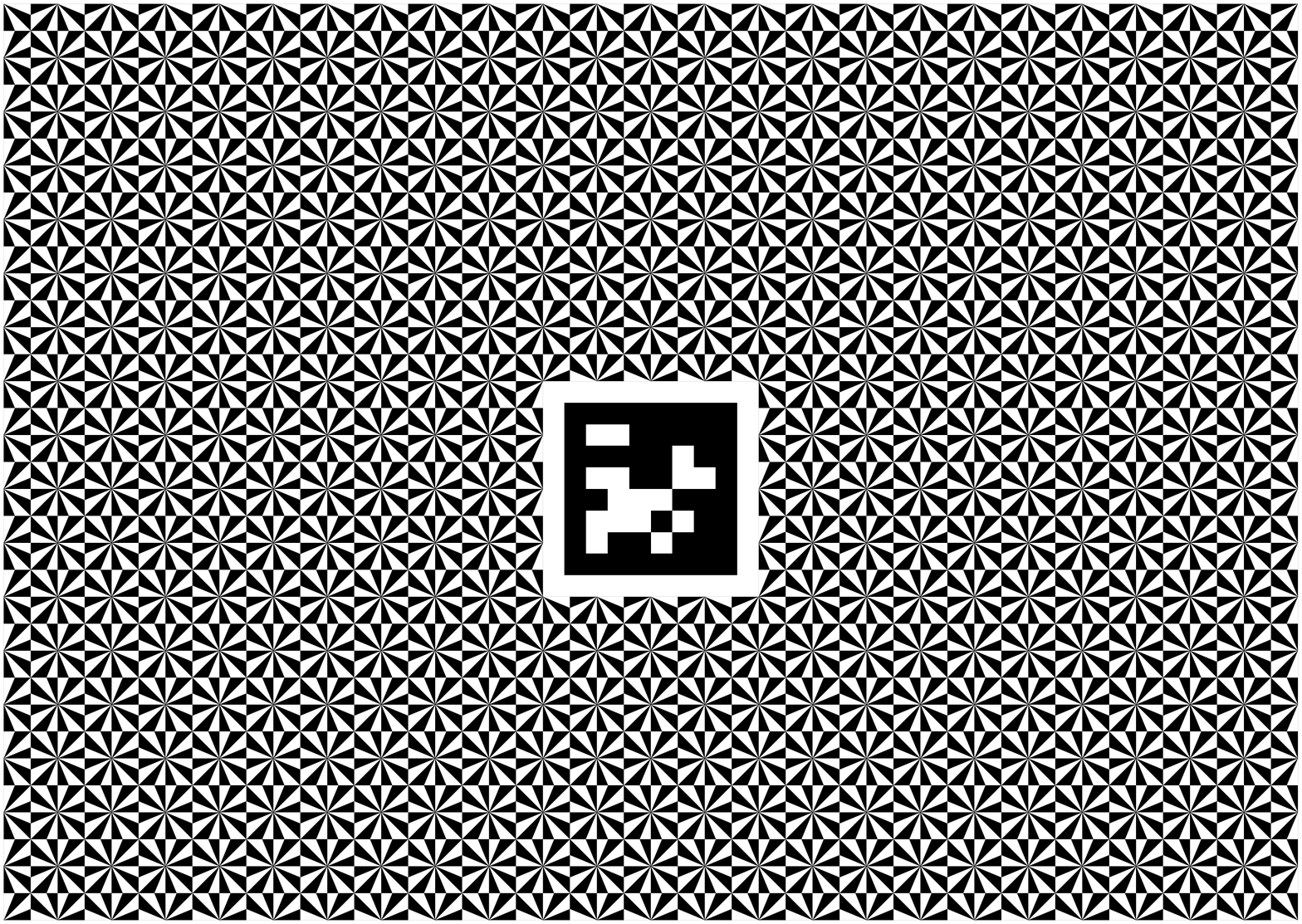}%
~~~~%
\includegraphics[trim={0 0 0 0},clip,height=0.35\linewidth]{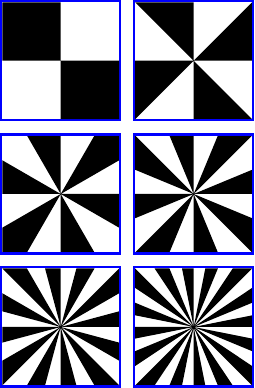}%
\end{center}
\vspace{-1em}
\caption{\textbf{Left:} Our (downsized) calibration pattern, %
allowing for unique localization using the AprilTag, and for very accurate feature detection using star-shaped feature points. %
Note that one should ideally adapt the density of the star squares to the resolution of the camera to be calibrated.
\textbf{Right:} Repeating pattern elements for different star segment counts. Top to bottom and left to right: 4 (checkerboard), 8, 12, 16, 24, 32 segments.
\vspace{-1.8em}
}
\label{fig:calibration_patterns}
\end{figure}

As mentioned in Sec.~\ref{sec:related_work},
dot patterns make it difficult for feature detection to be robust against %
distortion \cite{mallon2007pattern}.
We thus use patterns based on intersecting lines, such as checkerboards.
Checkerboards have several disadvantages however.
First, there is little image information around each corner to locate it:
Only the gradients of the two lines that intersect at the feature provide information.
As shown in \cite{ha2017deltille}, using 3 instead of 2 lines improves accuracy.
This %
raises the question whether the number of lines should be increased further.
Second, checkerboard corners change their appearance strongly when viewed under different rotations.
This may make feature detectors susceptible to yield differently biased results depending on the orientation of a feature, \eg, in the presence of chromatic aberration. %

To address these shortcomings, we propose to use star-based patterns (\cf Siemens stars, \eg, \cite{Reulke2006ISPRS}) as a generalization of checkerboards. 
Each feature in this type of pattern is the center of a star with a given number $s$ of alternating black and white segments.
For $s = 4$, the pattern corresponds to a checkerboard.
For $s = 6$, the features resemble those of the deltille pattern \cite{ha2017deltille} (while the feature arrangement differs from \cite{ha2017deltille}, however). %
We constrain the area of each star to a square and align these squares next to each other in a repeating pattern.
Additional corner features arise at the boundaries of these squares, which we however ignore, since their segment counts are in general lower than that of the feature in the center.
We also include an AprilTag \cite{wang2016iros} in the center of the pattern to facilitate its unambiguous localization (\cf \cite{ha2017deltille,beck2018generalized}).
See Fig.~\ref{fig:calibration_patterns} for an image of the pattern, and squares with different numbers of segments.
The number of segments needs to balance the amount of gradient information provided and the ability for the pattern to be resolved by the display or printing device and the camera; as justified in Sec.~\ref{sec:calibration_pattern_evaluation}, 
we use 16 segments.

The pattern can be simply printed onto a sheet of paper or displayed on a computer monitor.
If desired, multiple patterns can be used simultaneously, making it very easy to produce larger calibration geometries.
Strict planarity is not required, since we later perform full bundle adjustment including the calibration patterns' geometries.
However, we assume approximate planarity for initialization, and rigidity.

During data collection, we detect the features in real-time (\cf Sec.~\ref{sec:feature_extraction}) and visualize the pixels at which features have been detected.
This helps to provide detections in the whole image area. %
Image areas without detections either require regularization to fill in an estimated calibration, or need to be excluded from use. %
For global-shutter cameras, we record videos instead of images for faster recording.

\subsection{Feature Extraction}
\label{sec:feature_extraction}
Given an image of our 'star' calibration pattern (Fig.~\ref{fig:calibration_patterns}), we must accurately localize the star center features in the image.
We first detect them approximately and then refine the results.
For detection, we establish approximate local homographies between the image and the pattern, starting from the detected AprilTag corners.
Detected features add additional matched points and thus allow to expand the detection area.
For details, see the supplemental material.
In the following, we only detail the refinement, which determines the final accuracy, as this is the focus of this paper.

\begin{figure}
\begin{center}
\includegraphics[width=0.9\linewidth]{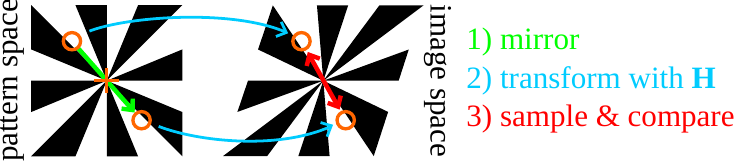}%
\end{center}
\vspace{-1em}
\caption{
Symmetry-based feature refinement:
Both a sample and its mirrored position (orange circles) are transformed from pattern to image space with homography $\mathbf{H}$.
$\mathbf{H}$ is optimized to minimize the differences between sampling both resulting positions.
\vspace{-1.8em}
}
\label{fig:symmetry_sketch}
\end{figure}

The refinement step receives an approximate feature location as input and needs to determine the feature's exact subpixel location.
To do so, we define a cost function based on symmetry (similar to the supplemental material of \cite{schoeps2019bad}), \cf Fig.~\ref{fig:symmetry_sketch}:
In pattern space, mirroring any point at a feature point must yield the same image intensity as the original point.
This is generally applicable to symmetrical patterns.

We define a local window for feature refinement which must include sufficient gradients, but should not include too much lens distortion. %
The optimum size depends on factors such as the blur from out-of-focus imaging, internal image processing in the camera, and clarity of the calibration pattern.
It should thus be suitably chosen for each situation; in this paper, we usually use $21{\times}21$ pixels.
It is not an issue if the window covers multiple 'stars' since the pattern is symmetric beyond a single star.
Within this window, we sample eight times as many random points as there are pixels in the window, in order to keep the variance due to random sampling low.
The initial feature detection (\cf supp.~PDF) provides a homography that locally maps between the pattern and image.
With this, we transform all $n$ random samples into pattern space, assuming the local window to be centered on the feature location.
A cost function $C_\text{sym}$ is then defined to compare points that are mirrored in pattern space:
\vspace{-1.6em}%
\setlength{\abovedisplayskip}{0.8em}%
\setlength{\belowdisplayskip}{0.4em}%
\setlength{\abovedisplayshortskip}{0.8em}%
\setlength{\belowdisplayshortskip}{0.4em}%
\begin{align}
C_\text{sym}(\mathbf{H}) = \sum_{i=1}^n  \big( I(\mathbf{H} (\mathbf{s}_i)) - I(\mathbf{H} (-\mathbf{s}_i)) \big)^2
\enspace.
\label{eq:symmetry_refinement}
\end{align}
\normalsize%
Here, $\mathbf{H}$ is the local homography estimate that brings pattern-space points into image space by homogeneous multiplication.
For each feature, we define it such that the origin in pattern space corresponds to the feature location.
$\mathbf{s}_i$ denotes the pattern-space location of sample $i$, and with the above origin definition, $-\mathbf{s}_i$ mirrors the sample at the feature.
$I$ is the image, accessed with bilinear interpolation.

We optimize $\mathbf{H}$ with the Levenberg-Marquardt method to minimize $C_\text{sym}$.
We fix the coefficient $\mathbf{H}_{2,2}$ to $1$ to obtain $8$ remaining parameters to optimize, corresponding to the $8$ degrees of freedom of the homography.
After convergence, we obtain the estimated feature location as $(\mathbf{H}_{0,2}, \mathbf{H}_{1,2})^T$.

The sample randomization reduces issues with bilinear interpolation:
For this type of interpolation, extrema of the interpolated values almost always appear at integer pixel locations.
This also makes cost functions defined on a regular grid of bilinearly-interpolated pixel values likely to have extrema there, which would introduce an unjustified prior on the probable subpixel feature locations.
Further, note that bilinear interpolation does not account for possible non-linearities in the camera's response function; however, these would be expected to only cause noise, not bias.

\subsection{Camera Model}
\label{sec:camera_model}
Accurate camera calibration requires a flexible model that avoids restricting the representable distortions. %
Storing a separate observation ray for each pixel, indicating where the observed light comes from, would be the most general model (assuming that a ray sufficiently approximates the origin directions). %
Such a model requires multiple feature observations for each pixel, or regularization, to be sufficiently constrained. %
Obtaining fully dense observations is very tedious with point features.
It would be more feasible with dense approaches \cite{rehder2017direct,rehder2017camera,hannemose2019superaccurate}, which we consider out of scope of this paper, and it would be possible with displayed patterns \cite{grossberg2001general,dunne2010efficient,bergamasco2013can,bergamasco2017parameter}, which we do not want to require.
We thus reduce the parameter count by storing observation rays in a regular grid in image space and interpolating between them %
(like \cite{rosebrock2012generic,rosebrock2012complete,beck2018generalized}).
This is appropriate for all cameras with smoothly varying observation directions. %

A non-central model, while potentially more accurate, may complicate the final application; images in general cannot be undistorted to the pinhole model, and algorithms designed for central cameras may need adaptation.
We %
consider both a central and a non-central model (\cf Fig.~\ref{fig:model_sketch}).

\PAR{Central camera model.}
For the central model, we store a unit-length observation direction at each grid point.
For un-projecting a given image pixel, these directions are interpolated as 3D points using a cubic B-Spline \cite{de1978practical} surface.
The interpolated point is then re-normalized to obtain the observation direction.
We also considered bicubic interpolation using Catmull-Rom splines \cite{catmull1974class}, however, the resulting surfaces tend to contain small wiggles as artifacts.

\PAR{Non-central camera model.}
When using the non-central model, each grid point stores both a unit-length direction and a 3D point $\mathbf{p}$ on the observation line.
In un-projection, both points are interpolated with a cubic B-Spline surface, and the direction is re-normalized afterwards. 
The result is a line passing through the interpolated 3D point with the computed direction.
Note that the interpolated lines may change if $\mathbf{p}$ is moved along the line.
Since, in contrast to the directions, there is no obvious normalization possibility for points $\mathbf{p}$, we keep this additional degree of freedom.

\PAR{Projection.}
The presented camera models define how to un-project pixels from the image to directions respectively lines in closed form.
For many applications and the later bundle adjustment step (\cf Sec.~\ref{sec:calibration}), the inverse is also required, \ie, projecting 3D points to pixels, which we find using an optimization process.
Note that this is different from many parametric models, which instead define projection in closed form and may require an optimization process for un-projection if they are not directly invertible.

To project a 3D point, we initialize the projected position in the center of the calibrated image area.
Then, similar to \cite{rosebrock2012generic}, we optimize it using the Levenberg-Marquardt method such that its un-projection matches the input point as closely as possible.
Pixel positions are constrained to the calibrated area, and we accept the converged result only if the final cost is below a very small threshold.
For speedup, if the same point was projected before with similar pose and intrinsics, the previous result can be used for initialization.

This approach worked for all tested cameras, as long as enough calibration data was recorded to constrain all grid parameters.
For cameras where the procedure might run into local minima, %
one could search over all observation directions / lines of the camera for those which match the input point best \cite{rosebrock2012generic}.
This also helps if one needs to know all projections in cases where points may project to multiple pixels, which is possible with both of our camera models.

\PAR{Performance.}
Tasks such as point (un)projection are low-level operations that may be performed very often in applications, %
thus their performance may be critical.
We thus shortly discuss the performance of %
our camera models.

For central cameras, images may be 'undistorted' to a different camera model, usually the pinhole model.
This transformation can be cached for calibrated cameras; once a lookup table for performing it is computed, the choice of the original camera model has no influence on the runtime anymore.
For high-field-of-view cameras, \eg, fisheye cameras, where undistortion to a pinhole model is impractical, one may 
use lookup tables 
from pixels to directions and vice versa. 
Thus, with an optimized implementation, there should be either zero or very little performance overhead when using generic models for central cameras.

For non-central cameras, image undistortion is not possible in general.
Un-projection can be computed directly (and cached in a lookup table for the whole image).
It should thus not be a performance concern.
However, projection may be slow; ideally, one would first use a fast approximate method (such as an approximate parametric model or lookup table) and then perform a few iterations of optimization to get an accurate result.
The performance of this may highly depend on the ability to quickly obtain good initial projection estimates for the concrete camera.

We think that given appropriate choice of grid resolution, the initial calibration should not take longer than 30 minutes up to sufficient accuracy on current consumer hardware.

\PAR{Parameter choice.} %
The grid resolution is the only parameter that must be set by the user.
The smallest interesting cell size is similar to the size of the feature refinement window (\cf Sec.~\ref{sec:feature_extraction}), since this window %
will generally 'blur' details with a kernel of this size.
Since we use $21{\times}21$ px or larger windows for feature extraction, we use grid resolutions down to $10$ px/cell, which we expect to leave almost no grid-based modeling error. %
If there is not enough data, the resolution should be limited to avoid overfitting.

\subsection{Calibration}
\label{sec:calibration}
Given images with extracted features, and the chosen central or non-central camera model, the model must be calibrated.
Our approach is to first initialize a per-pixel model on interpolated pattern matches using \cite{ramalingam2016unifying}.
Then we fit the final model to this, discard the interpolated matches, and obtain the final result with bundle adjustment.
See \cite{ramalingam2016unifying} and the supp.~material for details. %
In the following, we focus on the refinement step that is responsible for the final accuracy. %

\PAR{Bundle Adjustment.}
Bundle adjustment jointly refines the camera model parameters, image poses (potentially within a fixed multi-camera rig), and the 3D locations of the pattern features.
We optimize for the reprojection error, which is the standard cost function in bundle adjustment \cite{triggs1999bundle}:
\vspace{-0.5em}%
\setlength{\abovedisplayskip}{0.8em}%
\setlength{\belowdisplayskip}{0.4em}%
\setlength{\abovedisplayshortskip}{0.8em}%
\setlength{\belowdisplayshortskip}{0.4em}%
\begin{align}
C(\pi, \mathbf{M}, \mathbf{T}, \mathbf{p}) &= \sum_{c \in \mathcal{C}} \sum_{i \in \mathcal{I}_c} \sum_{o \in O_i} \rho(\mathbf{r}_{c,i,o}^T \mathbf{r}_{c,i,o})
\\
\mathbf{r}_{c,i,o} &= \pi_c(\mathbf{M}_c \mathbf{T}_i \mathbf{p}_o) - \mathbf{d}_{i,o} \nonumber
\vspace{-0.5em}
\end{align}
\normalsize%
Here, $\mathcal{C}$ denotes the set of all cameras, $\mathcal{I}_c$ the set of all images taken by camera $c$, and $O_i$ the feature observations in image $i$.
$\mathbf{p}_o$ is the 3D pattern point corresponding to observation $o$, and $\mathbf{d}_{i,o}$ the 2D detection of this point in image $i$.
$\mathbf{T}_i$ is the pose of image $i$ which transforms global 3D points into the local rig frame, and $\mathbf{M}_c$ %
transforms points within the local rig frame into camera $c$'s frame.
$\pi_c$ then projects the local point to camera $c$'s image using the current estimate of its calibration.
$\rho$ is a loss function on the squared residual; we use the robust Huber loss with parameter 1.

As is common, we optimize cost $C$ with the Levenberg-Marquardt method, and use local updates for orientations.
We also use local updates $(x_1, x_2)$ for the directions within the camera model grids: We compute two arbitrary, perpendicular tangents $\mathbf{t}_1, \mathbf{t}_2$ to each direction $\mathbf{g}$ and update it by adding a multiple of each tangent vector and then re-normalizing: $\frac{\mathbf{g} + x_1 \mathbf{t}_1 + x_2 \mathbf{t}_2}{\lVert \mathbf{g} + x_1 \mathbf{t}_1 + x_2 \mathbf{t}_2 \rVert}$.
Each grid point thus has a 2D 
update in the central model and a 5D 
update in the non-central one (two for the direction, and three for a 3D point on the line, \cf Sec.~\ref{sec:camera_model}). %
The projection $\pi$ involves an optimization process and the Inverse Function Theorem is not directly applicable. 
Thus, we use finite differences to compute the corresponding parts of the Jacobian.

The optimization process in our setting has more dimensions of Gauge freedom than for typical Bundle Adjustment problems, which we discuss in the supplemental material.
We experimented with Gauge fixing, but did not notice an advantage to explicitly fixing the Gauge directions; the addition of the Levenberg-Marquardt diagonal should already make the Hessian approximation invertible.

The Levenberg-Marquardt method compares the costs of different states to judge whether it makes progress.
However, we cannot always compute all residuals for each state.
During optimization, the projections of 3D points may enter and leave the calibrated image area, and since the camera model is only defined within this area, residuals cannot be computed for points that do not project into it.
If a residual is defined in one state but not in another, how should the states be compared in a fair way?
A naive solution would be to assign a constant value (\eg, zero) to a residual if it is invalid.
This causes state updates that make residuals turn invalid to be overrated, while
 updates that make residuals turn valid will be underrated.
As a result, the optimization could stall. %
Instead, we propose to compare states by summing the costs only for residuals which are valid in both states.
This way, cost comparisons are always fair; however, some residuals may not be taken into account.
Theoretically, this may lead to oscillation.
We however did not observe this in practice, and we believe that if it happens it will most likely be very close to the optimum, since otherwise the remaining residuals likely outweigh the few which change validity.
In such a case, it then seems safe to stop the optimization.

\section{Evaluation}
\label{sec:evaluation}
\begin{table}[t]
\scriptsize
\begin{center}
\begin{tabular}{@{\hspace{0.0em}}c@{\hspace{0.6em}}c@{\hspace{0.6em}}c@{\hspace{0.6em}}l@{\hspace{0.0em}}}
Label & Resolution & Field-of-view (FOV) & Description \\
\hline
D435-C & $1920 \times 1080$ & ca.~$70^\circ \times 42^\circ$ & Color camera of an Intel D435\\
D435-I & $1280 \times 800~~$ & ca.~$90^\circ \times 64^\circ$ & Infrared camera of an Intel D435\\
SC-C & $640 \times 480$ & ca.~$71^\circ \times 56^\circ$ & Color camera of a Structure Core\\
SC-I & $1216 \times 928~~$ & ca.~$57^\circ \times 45^\circ$ & Infrared camera of a Structure Core\\
\end{tabular}
\end{center}
\vspace{-1.2em}
\caption{
Specifications of the cameras used in the evaluation (more cameras are evaluated in the supplemental material).
FOV is measured horizontally and vertically at the center of the image.
\vspace{-1em}
}
\label{tab:evaluation_cameras}
\end{table}
Tab.~\ref{tab:evaluation_cameras} lists the cameras used for evaluation.
The camera labels from this table will be used throughout the evaluation. %

We evaluate the generic models against two parametric ones, both having 12 parameters, which is a high number for parametric models.
The first is the model implemented in OpenCV \cite{opencv_library} using all distortion terms.
The second is the Thin-Prism Fisheye model \cite{weng1992camera} with 3 radial distortion terms, which was used in a stereo benchmark~\cite{schoeps2017cvpr}.
We also consider a "Central Radial" model (similar to \cite{hartley2007parameter,tardif2008calibration,camposeco2015non}) based on the OpenCV model, adding the two thin-prism parameters from the Thin-Prism Fisheye model and replacing the radial term with a spline with many control points. %
With this, we evaluate how much improvement is obtained by better radial distortion modeling only.
Note that unfortunately, no implementations of complete generic calibration pipelines seemed to be available at the time of writing. %
This makes it hard to compare to other generic calibration pipelines; we released our approach as open source to change this.
In addition, since we aim to obtain the most accurate results possible, and since we avoid synthetic experiments as they often do not reflect realistic conditions, there is no ground truth to evaluate against. %
However, our main interest is in comparing to the commonly used parametric models to show why they should (if possible) be avoided.

\subsection{Calibration Pattern Evaluation}
\label{sec:calibration_pattern_evaluation}
\begin{figure}
\begin{center}
\includegraphics[trim={0 0 0 0},clip,width=1.0\linewidth]{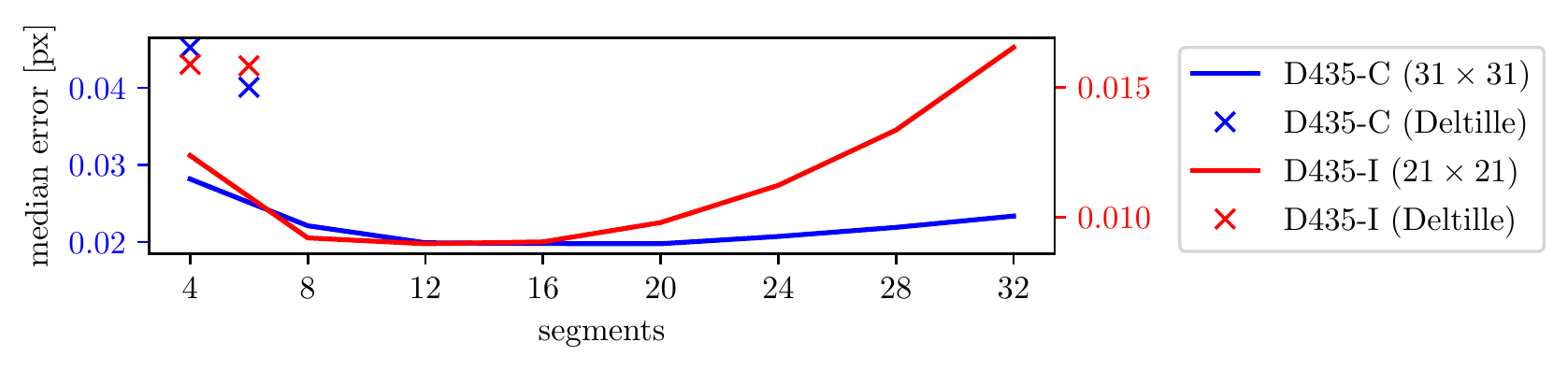}%
\end{center}
\vspace{-1.7em}
\caption{Median reprojection errors (y-axis) for calibrating cameras D435-C and D435-I with patterns having different numbers of star segments (x-axis).
The feature refinement window was $31{\times}31$ pixels for D435-C and $21{\times}21$ pixels for D435-I.
The Deltille results were obtained with the feature refinement from \cite{ha2017deltille}.
\vspace{-1.5em}
}
\label{fig:pattern_evaluation}
\end{figure}
We validate our choice of pattern (\cf Sec.~\ref{sec:data_collection}) by varying the number of star segments from 4 to 32 (\cf Fig.~\ref{fig:calibration_patterns}).
For the 4-segment checkerboard and 6-segment 'deltille' \cite{ha2017deltille} patterns, we also compare against the feature refinement from \cite{ha2017deltille}.
For each pattern variant, we record calibration images with the same camera from the same poses.
We do this by putting the camera on a tripod and showing the pattern on a monitor.
For each tripod pose that we use, we cycle through all evaluated patterns on the monitor to record a set of images with equal pose.
Since not all features are detected in all patterns, for fairness we only keep those feature detections which succeed for all pattern variants.

Since there is no ground truth for feature detections, we compare %
different patterns via the achieved reprojection errors.
We calibrate each set of images of a single pattern separately and compute its median reprojection error.
These results are plotted in Fig.~\ref{fig:pattern_evaluation}. %
Increasing the number of segments starting from 4, the accuracy is expected to improve first (since more gradients become available for feature refinement) and then worsen (since the monitor and camera cannot resolve the pattern anymore).
Both plots follow this expectation, with the best number of segments being 12 resp.~20. %
The experiment shows that neither the commonly used checkerboard pattern nor the 'deltille' pattern \cite{ha2017deltille} is optimal for either camera (given our feature refinement).
For this paper, %
we thus default to 16 segments as a good mean value.
The results of \cite{ha2017deltille} have higher error than ours for both the checkerboard and 'deltille' pattern.

\subsection{Feature Refinement Evaluation}
\label{sec:feature_refinement_evaluation}
\begin{figure}
\begin{center}
\includegraphics[trim={0 0 0 0},clip,width=0.99\linewidth]{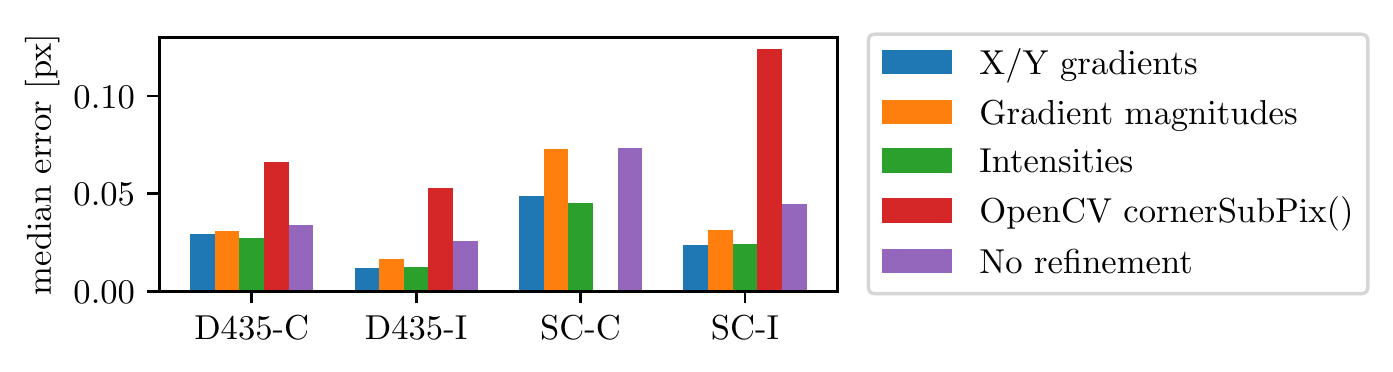}%
\end{center}
\vspace{-1.5em}
\caption{Median reproj.~error (y-axis) for calibrating the cameras on the x-axis with different feature refinement schemes (colors).
For SC-C, \texttt{cornerSubPix()} results were too inconsistent. %
\vspace{-1em}}
\label{fig:detector_evaluation}
\end{figure}
\begin{figure*}
\begin{center}
\footnotesize
\newcommand{\imagewidth}{0.128\linewidth}
\newcommand{\colwidth}{\hspace{0.2em}}
\newcommand{\figpic}[2]{%
\begin{tikzpicture}
    \node[anchor=south west,inner sep=0] (image) at (0,0) {\includegraphics[trim={12px 12px 12px 12px},clip,width=\imagewidth]{figures/model_evaluation_bias_free/#1}};
    \node [anchor=south east] at (image.south east) {\textcolor{white}{#2}};
\end{tikzpicture}
}
\begin{tabular}{@{\hspace{0em}}c@{\hspace{0.6em}}c@{\colwidth}c@{\colwidth}c@{\colwidth}c@{\colwidth}c@{\colwidth}c@{\colwidth}c@{\hspace{0em}}}
&%
OpenCV&%
Thin-Prism Fisheye&%
Central Radial&%
Central Generic&%
Central Generic&%
Central Generic&%
Noncentral Generic\\%
&%
(12 parameters)&%
(12 parameters)&%
(258 parameters)&%
ca.~30 px/cell&%
ca.~20 px/cell&%
ca.~10 px/cell&%
ca.~20 px/cell\\%
\rotatebox{90}{\begin{minipage}{4em}\centering D435-C\\(968\\images)\end{minipage}}&%
\includegraphics[trim={12px 12px 12px 12px},clip,width=\imagewidth]{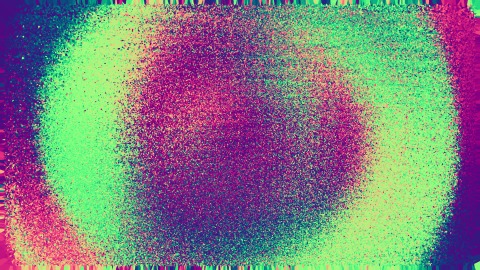}&%
\includegraphics[trim={12px 12px 12px 12px},clip,width=\imagewidth]{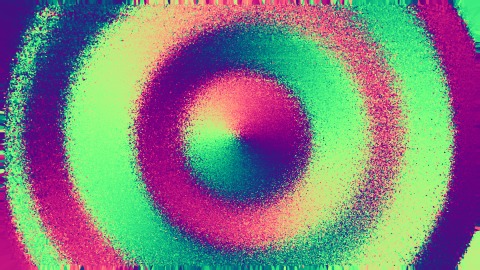}&%
\includegraphics[trim={12px 12px 12px 12px},clip,width=\imagewidth]{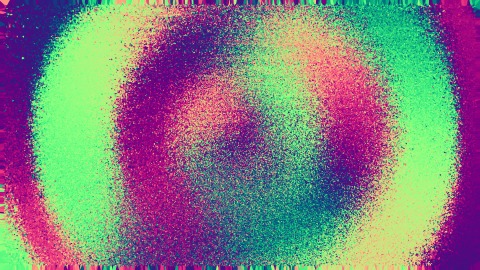}&%
\figpic{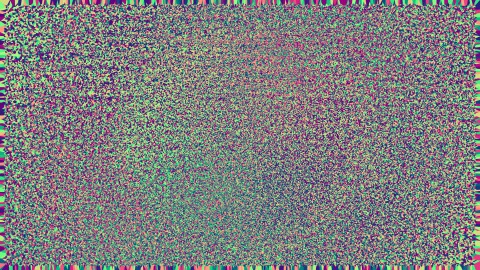}{4.8k}&%
\figpic{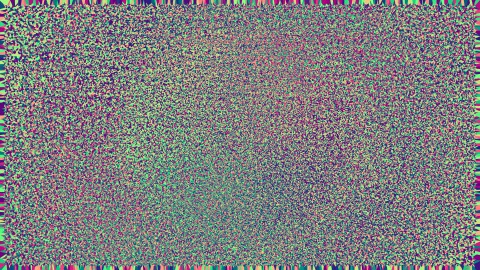}{10.4k}&%
\figpic{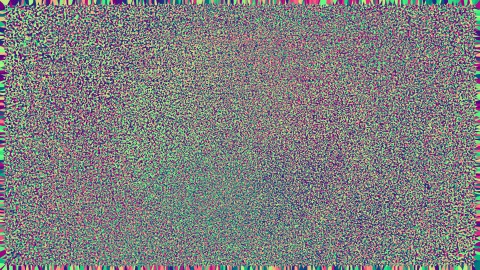}{40.9k}&%
\figpic{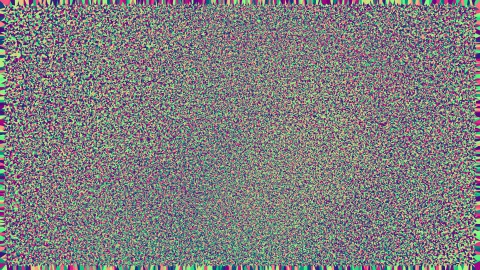}{25.9k}\\[-0.3em]%
Errors$^1$&%
\scriptsize
0.092 / 0.091 / 0.748&%
\scriptsize
0.163 / 0.161 / 1.379&%
\scriptsize
0.068 / 0.070 / 0.968&%
\scriptsize
0.030 / 0.039 / 0.264&%
\scriptsize
0.030 / 0.039 / 0.265&%
\scriptsize
0.029 / 0.040 / 0.252&%
\scriptsize
\textbf{0.024} / \textbf{0.032} / \textbf{0.184}\\%
\rotatebox{90}{\begin{minipage}{4.8em}\centering D435-I\\(1347\\images)\end{minipage}}&%
\includegraphics[trim={12px 12px 12px 12px},clip,width=\imagewidth]{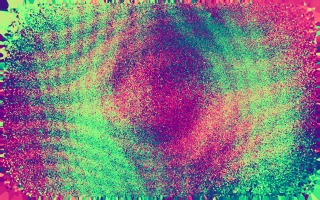}&%
\includegraphics[trim={12px 12px 12px 12px},clip,width=\imagewidth]{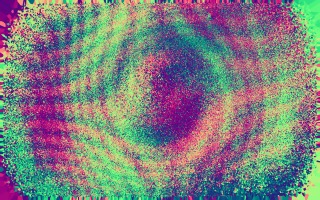}&%
\includegraphics[trim={12px 12px 12px 12px},clip,width=\imagewidth]{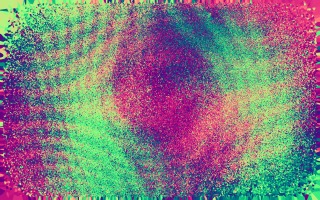}&%
\figpic{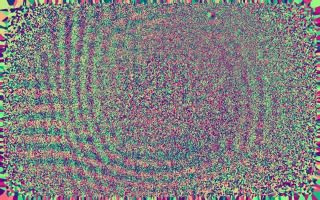}{2.5k}&%
\figpic{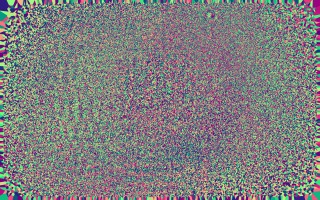}{5.3k}&%
\figpic{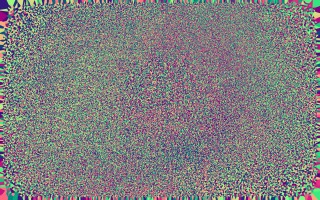}{20.5k}&%
\figpic{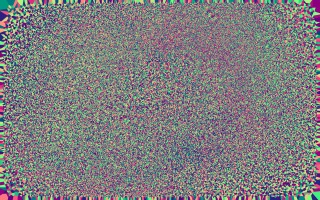}{13.3k}\\[-0.3em]%
Errors$^1$&%
\scriptsize
0.042 / 0.036 / 0.488&%
\scriptsize
0.032 / 0.026 / 0.365&%
\scriptsize
0.042 / 0.037 / 0.490&%
\scriptsize
0.023 / 0.018 / 0.199&%
\scriptsize
0.023 / 0.018 / 0.198&%
\scriptsize
0.023 / 0.018 / 0.189&%
\scriptsize
\textbf{0.022} / \textbf{0.017} / \textbf{0.179}\\%
\rotatebox{90}{\begin{minipage}{5.5em}\centering SC-C\\(1849\\images)\end{minipage}}&%
\includegraphics[trim={12px 12px 12px 12px},clip,width=\imagewidth]{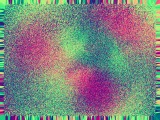}&%
\includegraphics[trim={12px 12px 12px 12px},clip,width=\imagewidth]{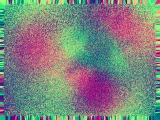}&%
\includegraphics[trim={12px 12px 12px 12px},clip,width=\imagewidth]{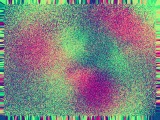}&%
\figpic{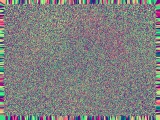}{0.7k}&%
\figpic{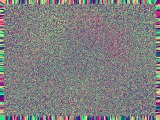}{1.5k}&%
\figpic{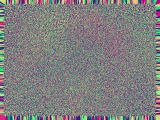}{5.9k}&%
\figpic{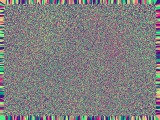}{3.8k}\\[-0.2em]%
Errors$^1$&%
\scriptsize
0.083 / 0.085 / 0.217&%
\scriptsize
0.083 / 0.084 / 0.215&%
\scriptsize
0.082 / 0.084 / 0.200&%
\scriptsize
0.069 / 0.072 / 0.055&%
\scriptsize
0.069 / 0.072 / 0.054&%
\scriptsize
0.068 / 0.072 / 0.053&%
\scriptsize
\textbf{0.065} / \textbf{0.069} / \textbf{0.040}\\%
\rotatebox{90}{\begin{minipage}{6em}\centering SC-I\\(2434\\images)\end{minipage}}&%
\includegraphics[trim={12px 12px 12px 12px},clip,width=\imagewidth]{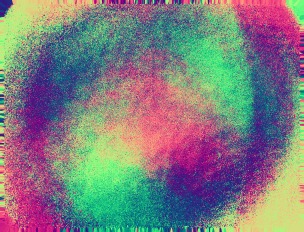}&%
\includegraphics[trim={12px 12px 12px 12px},clip,width=\imagewidth]{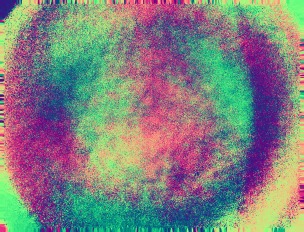}&%
\includegraphics[trim={12px 12px 12px 12px},clip,width=\imagewidth]{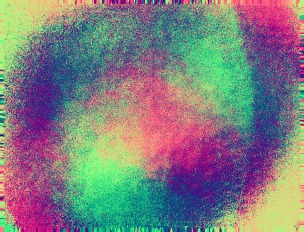}&%
\figpic{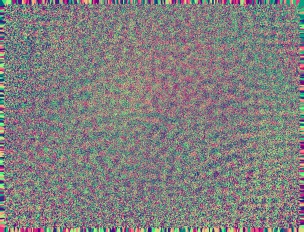}{2.5k}&%
\figpic{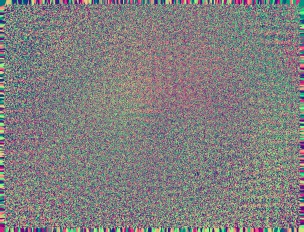}{5.6k}&%
\figpic{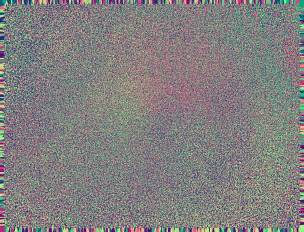}{21.8k}&%
\figpic{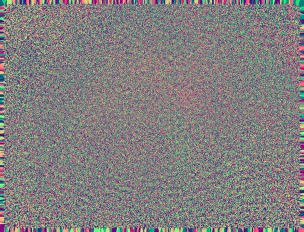}{14.0k}\\[-0.2em]%
Errors$^1$&%
\scriptsize
0.069 / 0.064 / 0.589&%
\scriptsize
0.053 / 0.046 / 0.440&%
\scriptsize
0.069 / 0.064 / 0.585&%
\scriptsize
0.035 / 0.030 / 0.133&%
\scriptsize
0.035 / 0.030 / 0.139&%
\scriptsize
0.034 / 0.030 / 0.137&%
\scriptsize
\textbf{0.030} / \textbf{0.026} / \textbf{0.120}\\%
\end{tabular}
\end{center}
\vspace{-0.7em}
\begin{minipage}[c]{0.09\linewidth}
\includegraphics[width=1.4cm]{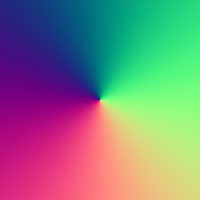}%
\vspace{-1.3em}
\end{minipage}%
\hfill%
\begin{minipage}[c]{0.91\linewidth}
\caption[]{
Directions (see legend on the left) of all reprojection errors for calibrating the camera given by the row with the model given by the column.
Each pixel shows the direction of the closest reprojection error %
from all images.
Ideally, the result is free from any systematic pattern.
Patterns indicate biased results arising from inability to model the true camera geometry.
Parameter counts for generic models are given in the images.
\hfill $^1$Median training error [px] / test error [px] / biasedness.
\vspace{-1.5em}
}\label{fig:model_eval_bias_free}
\end{minipage}
\end{figure*}
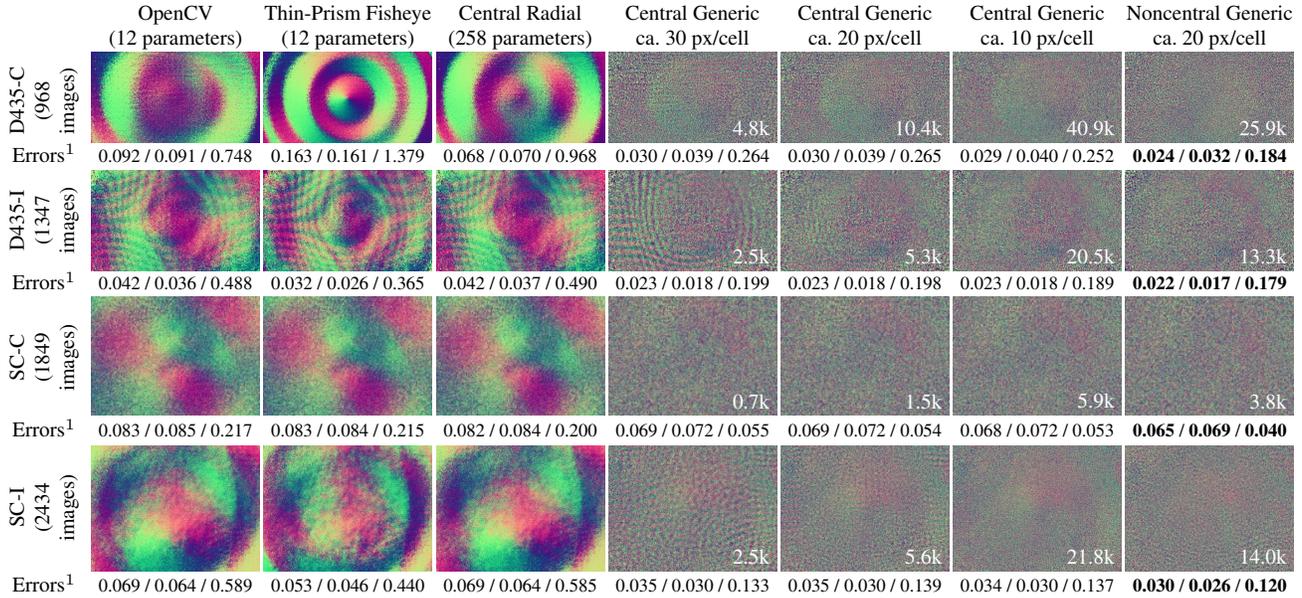
We compare several variants of our feature refinement (\cf Sec.~\ref{sec:feature_extraction}):
i) The original version of Eq.~\eqref{eq:symmetry_refinement},
and versions where we replace the raw intensity values by ii) gradient magnitudes, or iii) gradients (2-vectors).
In addition, we evaluate OpenCV's \cite{opencv_library} \texttt{cornerSubPix()} function, %
which implements \cite{forstner1987fast}.
In all cases, the initial feature positions for refinement are given by our feature detection scheme. %
For each camera, we take one calibration dataset, apply every feature refinement scheme on it, and compare the achieved median reprojection errors.
Similarly to the previous experiment, we only use features that are found by all methods.
The results are plotted in Fig.~\ref{fig:detector_evaluation}.
Intensities and X/Y gradients give the best results, with X/Y gradients performing slightly better for the monochrome cameras and intensities performing slightly better for the color cameras. %

\subsection{Validation of the Generic Model}
\label{sec:model_validation}
We validate that the generic models we use (\cf Sec.~\ref{sec:camera_model}) can calibrate cameras very accurately by verifying that they achieve bias-free calibrations:
The directions of the final reprojection errors should be random rather than having the same direction in parts of the image, which would indicate an inability of the model to fit the actual distortion in these areas.
 Fig.~\ref{fig:model_eval_bias_free} shows these directions for different cameras, calibrated with each tested model.
We also list the median reprojection errors, both on the calibration data and on a test set that was not used for calibration.
The latter is used to confirm that the models do not overfit.
As a metric of biasedness, we compute the KL-Divergence between the 2D normal distribution, and the empirical distribution of reprojection error vectors (scaled to have the same mean error norm), in each cell of a regular $50\times50$ grid placed on the image. 
We report the median value over these cells in Fig.~\ref{fig:model_eval_bias_free}.

The generic models achieve lower errors than the parametric ones throughout, while showing hardly any signs of overfitting.
This is expected, since -- given enough calibration images -- the whole image domain can be covered with training data, thus there will be no 'unknown' samples during test time.
Interestingly, the non-central model consistently performs best for all cameras %
in every metric, despite all of the cameras being standard near-pinhole cameras.

All parametric models show strong bias patterns in the error directions.
For some cameras, the generic models also show high-frequency patterns with lower grid resolutions that disappear with higher resolution.
These would be very hard to fit with any parametric model.
The central radial model only improves over the two parametric models for one camera, showing that improved radial distortion modeling alone is often not sufficient for significant improvement.

\subsection{Comparison of Different Models}
\label{sec:model_evaluation}
\begin{figure}
\begin{center}
\renewcommand{\tabcolsep}{0px}
\renewcommand{\arraystretch}{0}
\newcommand{\imgheight}{0.162\linewidth}
\begin{tabular}{cccc}
D435-C & D435-I & SC-C & SC-I\\[0.3em]
\includegraphics[height=\imgheight]{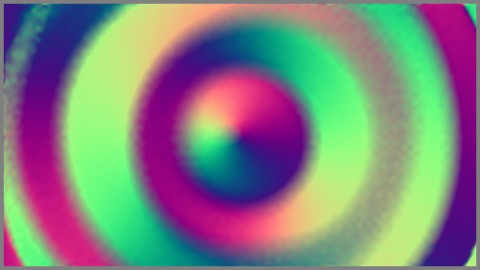}&%
\includegraphics[height=\imgheight]{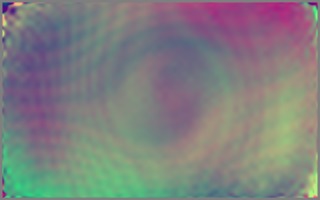}&%
\includegraphics[height=\imgheight]{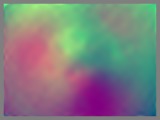}&%
\includegraphics[height=\imgheight]{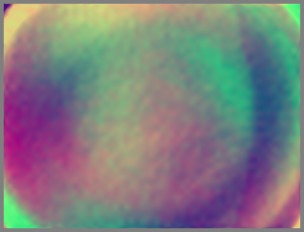}\\
\includegraphics[height=\imgheight]{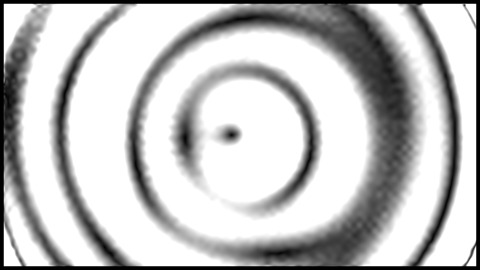}&%
\includegraphics[height=\imgheight]{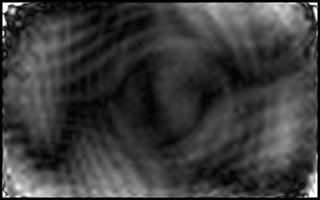}&%
\includegraphics[height=\imgheight]{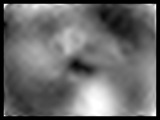}&%
\includegraphics[height=\imgheight]{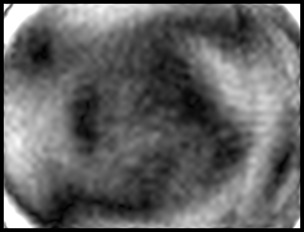}\\%
\end{tabular}
\end{center}
\vspace{-0.6em}
\caption{Differences between calibrations with the central-generic model and fitted Thin-Prism Fisheye calibrations, measured as reprojection errors.
\textbf{Top:} Medium gray corresponds to zero error, %
while saturated colors as in Fig.~\ref{fig:model_eval_bias_free} correspond to 0.2 pixels difference.
\textbf{Bottom:} Alternative visualization showing the error magnitude only, with black for zero error and white for 0.2 pixels error.
\vspace{-3em}}
\label{fig:model_comparison}
\end{figure}
We now take a closer look at the differences between accurate calibrations and calibrations obtained with typical parametric models.
We fit the Thin-Prism Fisheye model to our calibrations, optimizing the model parameters to minimize the two models' deviations in the observation directions per-pixel.
At the same time, we optimize for a 3D rotation applied to the observation directions of one model, since consistent rotation of all image poses for a camera can be viewed as part of the intrinsic calibration.
After convergence, we visualize the remaining differences in the observation directions. %
While these visualizations will naturally be similar to those of Fig.~\ref{fig:model_eval_bias_free} given our model is very accurate, we can avoid showing the feature detection noise here.
Here, we visualize both direction and magnitude of the differences, while Fig.~\ref{fig:model_eval_bias_free} only visualizes directions.
The results are shown in Fig.~\ref{fig:model_comparison}, and confirm that the models differ in ways that would be difficult to model with standard parametric models.
Depending on the camera and image area, the reprojection differences are commonly up to 0.2 pixels, or even higher for the high-resolution camera D435-C.

\subsection{Example Application: Stereo Depth Estimation}
\label{sec:application_example}
\begin{figure}
\begin{center}
\renewcommand{\tabcolsep}{0px}
\renewcommand{\arraystretch}{0}
\begin{tabular}{cc}
\includegraphics[height=0.33\linewidth]{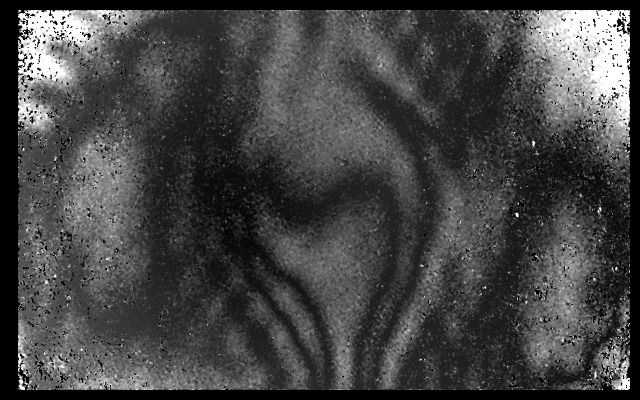}&%
\includegraphics[height=0.33\linewidth]{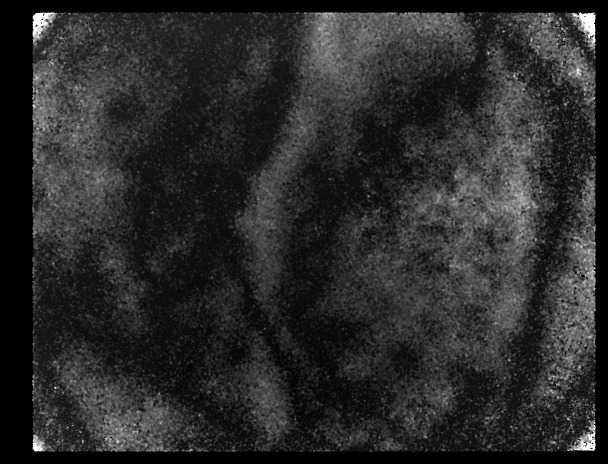}\\[-1.0em]%
\textcolor{white}{D435} & \textcolor{white}{Structure Core}\\[0.4em]
\end{tabular}
\end{center}
\vspace{-0.6em}
\caption{
Distances (black: $0cm$, white: $1cm$) between corresponding points estimated by dense stereo with a generic and a parametric calibration, at roughly 2 meters depth. %
\vspace{-1.7em}
}
\label{fig:stereo_relative_distances}
\end{figure}
So far, we showed that generic models yield better calibrations than common parametric ones.
However, the differences might appear small, and it might thus be unclear how valuable they are in practice.
Thus, we now look at the role of small calibration errors in example applications. %

Concretely, we consider dense depth estimation for the Intel D435 and Occipital Structure Core active stereo cameras.
These devices contain infrared camera pairs with a relatively small baseline, as well as an infrared projector that provides texture for stereo matching.
The projector behaves like an external light and thus does not need to be calibrated; only the calibration of the stereo cameras is relevant.

Based on the previous experiments, we make the conservative assumption that the calibration error for parametric models will be at least $0.05$ pixels in many parts of the image.
Errors in both stereo images may add up or cancel each other out depending on their directions.
A reasonable assumption is that the calibration error will lead to a disparity error of similar magnitude.
Note that for typical stereo systems, the stereo matching error for easy-to-match surfaces may be assumed to be as low as $0.1$ pixels \cite{schoeps2019bad}; in this case, the calibration error may even come close to the level of noise.
The well-known relation between disparity $x$ and depth $d$ is: $d = \frac{bf}{x}$, with baseline $b$ and focal length $f$.
Let us consider $b = 5 cm$ and $f = 650 px$ (roughly matching the D435).
For $d = 2 m$ for example, a disparity error of $\rpm 0.05 px$ results in a depth error of about $0.6 cm$. %
This error grows quadratically with depth, and since it stays constant over time, it acts as a \emph{bias} that will not easily average out.

For empirical validation, we calibrate the stereo pairs of a D435 and a Structure Core device with both the central-generic and the Thin-Prism Fisheye model (which fits the D435-I and SC-I cameras better than the OpenCV model, see Fig.~\ref{fig:model_eval_bias_free}).
With each device, we recorded a stereo image of a roughly planar wall in approx.~$2 m$ distance and estimated a depth image for the left camera with both calibrations.
Standard PatchMatch Stereo \cite{bleyer2011patchmatch} with Zero-Mean Normalized Cross Correlation costs works well given the actively projected texture.
The resulting point clouds were aligned with a similarity transform with the Umeyama method \cite{umeyama1991least}, since the different calibrations may introduce scale and orientation differences. %
Fig.~\ref{fig:stereo_relative_distances} shows the distances of corresponding points in the aligned clouds.
Depending on the image area, the error is often about half a centimeter, and goes up to more than 1 cm for both cameras.
This matches the theoretical result from above
well and shows that one should avoid such a bias for accurate results.

\subsection{Example Application: Camera Pose Estimation}
\label{sec:application_example_localization}
To provide a broader picture, we also consider camera pose estimation as an application.
For this experiment, we treat the central-generic calibration as ground truth and sample 15 random pixel locations in the image.
We un-project each pixel to a random distance to the camera from 1.5 to 2.5 meters.
Then we change to the Thin-Prism Fisheye model and localize the camera with the 2D-3D correspondences defined above.
The median error in the estimated camera centers is 2.15 mm for D435-C, 0.25 mm for D435-I, 1.80 mm for SC-C, and 0.76 mm for SC-I.

Such errors may accumulate during visual odometry or SLAM.
To test this, we use Colmap \cite{schoenberger2016sfm} on several videos and bundle-adjust the resulting sparse reconstructions both with our Thin-Prism-Fisheye and non-central generic calibrations.
For each reconstruction pair, we align the scale and the initial camera poses of the video, and compute the resulting relative translation error at the final image compared to the trajectory length.
For camera D435-I, we obtain $0.2\% \pm 0.1\%$ error, while for SC-C, we get $2.9\% \pm 2.3\%$.
These errors strongly depend on the camera, reconstruction system, scene, and trajectory, so our results only represent examples.
However, they clearly show that even small calibration improvements can be significant in practice.

\vspace{-8pt}
\section{Conclusion}
\vspace{-8pt}
We proposed a generic camera calibration pipeline which focuses on accuracy while being easy to use.
It achieves virtually bias-free results in contrast to using parametric models;
for all tested 
cameras, the non-central generic model performs best.
We also showed that even small calibration improvements can be valuable in practice, since they avoid biases that may be hard to average out.

Thus, we believe that generic models should replace parametric ones as the default solution for camera calibration.
If a central model is used, this might not even introduce a performance penalty, since the runtime performance of image undistortion via lookup does not depend on the original model. %
We facilitate the use of generic models by releasing our calibration pipeline as open source.
However, generic models might not be suitable for all use cases, in particular if the performance of projection to distorted images is crucial, if self-calibration is required, or if not enough data for dense calibration is available.
{\small
\PAR{Acknowledgements.} 
Thomas Sch\"{o}ps was partially supported by a Google PhD Fellowship. 
Viktor Larsson was supported by the ETH Zurich Postdoctoral Fellowship program. 
This work was supported by the Swedish Foundation for Strategic Research (Semantic Mapping and Visual Navigation for Smart Robots). 
}


\title{Supplemental Material for:\\Why Having 10,000 Parameters in Your Camera Model is Better Than Twelve}

\author{Thomas Sch\"ops$^1$
\qquad
Viktor Larsson$^1$
\qquad
Marc Pollefeys$^{1,2}$
\qquad
Torsten Sattler$^3$\\
$^1$%
Department of Computer Science, ETH Z\"urich\\%
$^2$Microsoft, Zurich%
\quad
$^3$Chalmers University of Technology\\
}

\maketitle

In this supplemental material, we present additional information that did not fit into the paper for space reasons.
In Sec.~\ref{sec:feature_detection}, we present the feature detection process that we use to find the approximate locations of star center features in images of our calibration pattern.
In Sec.~\ref{sec:calibration_initialization}, we present the initialization process for camera calibration.
In Sec.~\ref{sec:gauge_freedom}, we discuss additional details of our bundle adjustment.
In Sec.~\ref{sec:model_validation_supp}, we show results of the camera model validation experiment for more cameras.
In Sec.~\ref{sec:application_example_localization_supp}, we present more details of the results of the Structure-from-Motion experiment that tests the impact of different camera models in an application context.

\section{Feature detection}
\label{sec:feature_detection}
First, we find all AprilTags in the image using the AprilTag library \cite{wang2016iros}.
The four corners of each detected AprilTag provide correspondences between the known calibration pattern and the image.
We use these to compute a homography that approximately maps points on the pattern into the image.
This will only be perfectly accurate for pinhole cameras and planar patterns. 
However, in general it will be locally appropriate next to the correspondences that were used to define the homography.
With this, we roughly estimate the positions of all star features that are directly adjacent to an AprilTag.
Each feature location is then refined and validated with the refinement process detailed below.
After refinement, the final feature positions provide additional correspondences between the pattern and the image.
For each newly detected and refined feature location, we compute a new local homography from the correspondences that are closest to it in order to predict its remaining adjacent feature locations that have not been detected yet.
We then repeat the process of feature refinement, local homography estimation, and growing, until no additional feature can be detected. %

The initial feature locations predicted by the above procedure can be relatively inaccurate.
Thus, we first apply a well-converging matching process based on the known pattern appearance to refine and validate the feature predictions, before we apply an accurate refinement step with a smaller convergence region afterwards.

\PAR{Matching-based refinement and validation.}
\begin{figure}
\begin{center}
\includegraphics[width=1.0\linewidth]{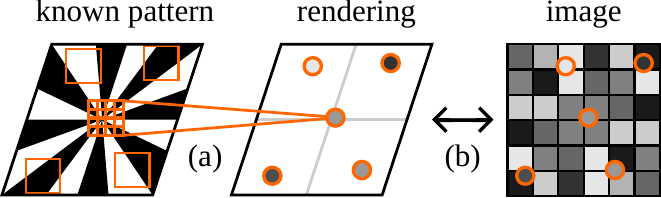}%
\end{center}
\caption{
Sketch of matching-based feature refinement.
(a) Based on an estimate for a local homography between the pattern and the image, the known pattern is rendered with supersampling (subpixels illustrated for center sample only for clarity).
The result are rendered grayscale samples shown in the center.
(b) The samples are (rigidly) moved within the image to find the best matching feature position, while accounting for affine brightness differences.
}
\label{fig:matching_sketch}
\end{figure}
The goal of matching-based refinement is to improve an initial feature position estimate and reject wrong estimates.
As input, the process described above yields an initial rough estimate of the feature position and a homography that locally approximates the mapping between the known calibration pattern and the image.
The matching process uses the homography to render a synthetic image of the pattern in a small window around the feature position.
Then it locally optimizes the alignment between this rendering and the actual image.
This is illustrated in Fig.~\ref{fig:matching_sketch}.

In detail, we define a local window for refinement, for which we mostly use $21{\times}21$ pixels.
The optimum size depends on many factors such as the blur introduced by out-of-focus imaging, internal image processing in the camera, and clarity of the calibration pattern.
It is thus a parameter that should be tuned to the specific situation.
Within this window, we sample as many random points as there are pixels in the window.
We assume that the window is centered on the feature location, and given the local homography estimate, we determine the intensity that would be expected for a perfect pattern observation at each sample location.
We use 16x supersampling to more accurately predict the intensities.

Next, we match this rendering with the actual image observation by optimizing for a 2-dimensional translational shift $\mathbf{x}$ of all samples in the image.
In addition, since the black and white parts of the pattern will rarely be perfectly black and white under real imaging conditions, we optimize for an affine brightness transform to bring the sample intensities and image intensities into correspondence.
In total, we thus optimize for four parameters with the following cost function:
\begin{align}
C_\text{match}(\mathbf{x}, f, b) = \sum_i^n \left( f \cdot p_i(\mathbf{x}) + b - q_i \right)^2  \enspace,
\end{align}
where $f$ and $b$ are the affine factor and bias, $p_i(\mathbf{x})$ is the bilinearly interpolated image intensity at the sample position $i$ with the current translation shift $\mathbf{x}$, and $q_i$ is the rendered intensity of sample $i$.
Given the initial translation offset $\mathbf{x} = \mathbf{0}$, we can initialize $f$ and $b$ directly by minimizing $C_\text{match}$.
Setting both partial derivatives to zero eventually yields (dropping $i$ from the notation for brevity):
\begin{align}
f = \dfrac{\sum (q p) - \frac{1}{n} \sum p \sum q}{\sum (p p) - \frac{1}{n} (\sum p)^2} \enspace , \enspace
b = \dfrac{\sum q - f \sum p}{n} \enspace.
\end{align}
Subsequently, we optimize all four parameters with the Levenberg-Marquardt method.
The factor parameter $f$ is not constrained in this optimization and may become negative, indicating that we have likely found a part of the image that looks more like the inverse of a feature than the feature itself.
While this may appear like a deficiency, since pushing the parameter to be positive might have nudged the translation $\mathbf{x}$ to go towards the proper feature instead, we can actually use it to our advantage, as we can use the condition of $f$ to be positive as a means to reject outliers.
This allows us to obtain virtually outlier-free detections.

Note that for performance reasons, in this part of the refinement we do not optimize for the full homography that is used at the start to render the pattern prediction.
This changes in the following symmetry-based refinement step, which is described in Sec.~\ref{sec:feature_extraction} in the paper.

\begin{table*}[t]
\small
\begin{center}
\begin{tabular}{lcccl}
      & Used       & Field-of-view (FOV) &      &             \\
Label & resolution & (approximate)       & Type & Description \\
\hline
D435-C & $1920 \times 1080$ & $70^\circ \times 42^\circ$ & RGB & Color camera of an Intel RealSense D435\\
D435-I & $1280 \times 800~~$ & $90^\circ \times 64^\circ$ & Mono & Infrared camera of an Intel RealSense D435\\
SC-C & $640 \times 480$ & $71^\circ \times 56^\circ$ & RGB & Color camera of an Occipital Structure Core (color version)\\
SC-I & $1216 \times 928~~$ & $57^\circ \times 45^\circ$ & Mono & Infrared camera of an Occipital Structure Core (color version)\\
Tango & $640 \times 480$ & $131^\circ \times 99^\circ~~$ & Mono & Fisheye camera of a Google Tango Development Kit Tablet\\
FPGA & $748 \times 468$ & $111^\circ \times 66^\circ~~$ & Mono & Camera attached to an FPGA\\
GoPro & $3000 \times 2250$ & $123^\circ \times 95^\circ~~$ & RGB & GoPro Hero4 Silver action camera\\
HTC One M9 & $3840 \times 2688$ & $64^\circ \times 47^\circ$ & RGB & Main (back) camera of an HTC One M9 smartphone\\
\end{tabular}
\end{center}
\vspace{-1em}
\caption{
Specifications of the cameras used in the evaluation: The resolution at which we used them, and the approximate field-of-view, which is measured horizontally and vertically at the center of the image.
SC-I provides images that are pre-undistorted by the camera.
\vspace{-1em}
}
\label{tab:evaluation_cameras_supp}
\end{table*}

\section{Calibration initialization}
\label{sec:calibration_initialization}
In this section, we describe how we obtain an initial camera calibration that is later refined with bundle adjustment, as described in the paper.

For each image, we first interpolate the feature detections over the whole pattern area for initialization purposes, as in \cite{ramalingam2016unifying}.
This is important to get feature detections at equal pixels in different images, which is required for the relative pose initialization approach \cite{ramalingam2016unifying} that is used later.
Since we know that the calibration pattern is approximately planar, 
we can use a homography to approximately map between the pattern and the image (neglecting lens distortion).
Each square of four adjacent feature positions is used to define a homography, which we use for mapping within this square.
This allows to obtain dense approximate pattern coordinates for all image pixels at which the pattern is visible.
These approximately interpolated matches are only used for initialization, not for the subsequent refinement.

We then randomly sample up to 500 image triples from the set of all available images.
We score each triple based on the number of pixels that have a dense feature observation in each of the three images. %
The image triple with the highest number of such pixels is used for initialization. %

Since all tested cameras were near-central, we always assume a central camera during initialization (and switch to the non-central model later if requested).
We thus use the linear pose solver for central cameras and planar calibration targets from \cite{ramalingam2016unifying}. %
For each image pixel which has an (interpolated) feature observation in each of the three images chosen above, the corresponding known 3D point on the calibration pattern is inserted into a 3D point cloud for each image.
The initialization approach \cite{ramalingam2016unifying} is based on the fact that for a given observation, the corresponding points in the three point clouds must lie on the same line in 3D space.
It solves for an estimate of the relative pose of the three initial images, and the position of the camera's optical center.
This allows to project the pattern into image space for each pixel with a matched pattern coordinate, which initializes the observation direction for these pixels.

Next, we extend the calibration by localizing additional images using the calibration obtained so far with standard techniques \cite{kneip2014opengv}.
Each localized image can be used to project the calibration pattern into image space, as above, and extend the calibrated image region.
For pixels that already have an estimate of their observation direction, we use the average of all directions to increase robustness.
If more than one calibration pattern is used, we can determine the relative pose between the targets from images in which multiple targets are visible.
We localize as many images as possible with the above scheme.

As a result, we obtain a per-pixel camera model which stores an observation direction for each initialized pixel. %
We then fit the final camera model to this initialization by first setting the direction of each grid point in the final model to the direction of its corresponding pixel in the per-pixel model.
If the pixel does not have a direction estimate, we guess it based on the directions of its neighbors.
Finally, using a non-linear optimization process with the Levenberg-Marquardt method, we optimize the model parameters such that the resulting observation directions for each pixel match the per-pixel initialization as closely as possible.

Due to the size of the local window in feature refinement (\cf the section on feature extraction in the paper), features cannot be detected close to the image borders (since the window would leave the image).
We thus restrict the fitted model to the axis-aligned bounding rectangle of the feature observations.

\section{Bundle adjustment details}
\label{sec:gauge_freedom}
\PAR{Gauge freedom.}
As mentioned in the paper, in our setting there are more dimensions of Gauge freedom than for typical Bundle Adjustment problems.
Note that we do not use any scaling information for the pattern(s) during bundle adjustment, but scale its result once as a post-processing step based on the known physical pattern size.
For the central camera model, the Gauge freedom dimensions are thus: 3 for global translation, 3 for global rotation, 1 for global scaling, and 3 for rotating all camera poses in one direction while rotating all calibrated observation directions in the opposite direction.
For the non-central camera model, the Gauge freedom dimensions are those listed for the central model and additionally 3 for moving all camera poses in one direction while moving all calibrated lines in the opposite direction.
Furthermore, if the calibrated lines are (nearly) parallel, there can be more directions, since then the cost will be invariant to changes of the 3D line origin points within the lines.

\PAR{Calibration data bias.}
For parametric models, whose parameters affect the whole image area, different densities in detected features may introduce a bias, since the camera model will be optimized to fit better to areas where there are more feature detections than to areas where there are less.
Note that this is a reason for image corners typically being modeled badly with these models, since there typically are very few observations in the corners compared to the rest of the image, and the corners are at the end of the range of radius values that are relevant for radial distortion.

For our generic models, while there is some dependence among different image areas due to interpolation within the grid, they are mostly independent.
Thus, this kind of calibration data bias should not be a concern for our models.
However, unless using regularization, all parts of the image need to contain feature detections to be well-constrained (which is again most difficult for the image corners).

\section{Camera model validation}
\label{sec:model_validation_supp}
\begin{figure*}
\begin{center}
\footnotesize
\newcommand{\imagewidth}{0.128\linewidth}
\newcommand{\colwidth}{\hspace{0.2em}}
\newcommand{\figpic}[2]{%
\begin{tikzpicture}
    \node[anchor=south west,inner sep=0] (image) at (0,0) {\includegraphics[trim={12px 12px 12px 12px},clip,width=\imagewidth]{figures/model_evaluation_bias_free/#1}};
    \node [anchor=south east] at (image.south east) {\textcolor{white}{#2}};
\end{tikzpicture}
}
\begin{tabular}{@{\hspace{0em}}c@{\hspace{0.6em}}c@{\colwidth}c@{\colwidth}c@{\colwidth}c@{\colwidth}c@{\colwidth}c@{\colwidth}c@{\hspace{0em}}}
&%
OpenCV&%
Thin-Prism Fisheye&%
Central Radial&%
Central Generic&%
Central Generic&%
Central Generic&%
Noncentral Generic\\%
&%
(12 parameters)&%
(12 parameters)&%
(258 parameters)&%
ca.~30 px/cell&%
ca.~20 px/cell&%
ca.~10 px/cell&%
ca.~20 px/cell\\%
\rotatebox{90}{\begin{minipage}{4em}\centering D435-C\\(968\\images)\end{minipage}}&%
\includegraphics[trim={12px 12px 12px 12px},clip,width=\imagewidth]{figures/model_evaluation_bias_free/D435_color/result_15px_central_opencv.jpg}&%
\includegraphics[trim={12px 12px 12px 12px},clip,width=\imagewidth]{figures/model_evaluation_bias_free/D435_color/result_15px_central_thin_prism_fisheye.jpg}&%
\includegraphics[trim={12px 12px 12px 12px},clip,width=\imagewidth]{figures/model_evaluation_bias_free/D435_color/result_15px_central_radial.jpg}&%
\figpic{D435_color/result_15px_central_generic_30.jpg}{4.8k}&%
\figpic{D435_color/result_15px_central_generic_20.jpg}{10.4k}&%
\figpic{D435_color/result_15px_central_generic_10.jpg}{40.9k}&%
\figpic{D435_color/result_15px_noncentral_generic_20.jpg}{25.9k}\\[-0.3em]%
Errors$^1$&%
\scriptsize
0.092 / 0.091 / 0.748&%
\scriptsize
0.163 / 0.161 / 1.379&%
\scriptsize
0.068 / 0.070 / 0.968&%
\scriptsize
0.030 / 0.039 / 0.264&%
\scriptsize
0.030 / 0.039 / 0.265&%
\scriptsize
0.029 / 0.040 / 0.252&%
\scriptsize
\textbf{0.024} / \textbf{0.032} / \textbf{0.184}\\%
\rotatebox{90}{\begin{minipage}{4.8em}\centering D435-I\\(1347\\images)\end{minipage}}&%
\includegraphics[trim={12px 12px 12px 12px},clip,width=\imagewidth]{figures/model_evaluation_bias_free/D435_infrared1/result_10px_central_opencv.jpg}&%
\includegraphics[trim={12px 12px 12px 12px},clip,width=\imagewidth]{figures/model_evaluation_bias_free/D435_infrared1/result_10px_central_thin_prism_fisheye.jpg}&%
\includegraphics[trim={12px 12px 12px 12px},clip,width=\imagewidth]{figures/model_evaluation_bias_free/D435_infrared1/result_10px_central_radial.jpg}&%
\figpic{D435_infrared1/result_10px_central_generic_30.jpg}{2.5k}&%
\figpic{D435_infrared1/result_10px_central_generic_20.jpg}{5.3k}&%
\figpic{D435_infrared1/result_10px_central_generic_10.jpg}{20.5k}&%
\figpic{D435_infrared1/result_10px_noncentral_generic_20.jpg}{13.3k}\\[-0.3em]%
Errors$^1$&%
\scriptsize
0.042 / 0.036 / 0.488&%
\scriptsize
0.032 / 0.026 / 0.365&%
\scriptsize
0.042 / 0.037 / 0.490&%
\scriptsize
0.023 / 0.018 / 0.199&%
\scriptsize
0.023 / 0.018 / 0.198&%
\scriptsize
0.023 / 0.018 / 0.189&%
\scriptsize
\textbf{0.022} / \textbf{0.017} / \textbf{0.179}\\%
\rotatebox{90}{\begin{minipage}{5.5em}\centering SC-C\\(1849\\images)\end{minipage}}&%
\includegraphics[trim={12px 12px 12px 12px},clip,width=\imagewidth]{figures/model_evaluation_bias_free/Core_color/result_15px_central_opencv.jpg}&%
\includegraphics[trim={12px 12px 12px 12px},clip,width=\imagewidth]{figures/model_evaluation_bias_free/Core_color/result_15px_central_thin_prism_fisheye.jpg}&%
\includegraphics[trim={12px 12px 12px 12px},clip,width=\imagewidth]{figures/model_evaluation_bias_free/Core_color/result_15px_central_radial.jpg}&%
\figpic{Core_color/result_15px_central_generic_30.jpg}{0.7k}&%
\figpic{Core_color/result_15px_central_generic_20.jpg}{1.5k}&%
\figpic{Core_color/result_15px_central_generic_10.jpg}{5.9k}&%
\figpic{Core_color/result_15px_noncentral_generic_20.jpg}{3.8k}\\[-0.2em]%
Errors$^1$&%
\scriptsize
0.083 / 0.085 / 0.217&%
\scriptsize
0.083 / 0.084 / 0.215&%
\scriptsize
0.082 / 0.084 / 0.200&%
\scriptsize
0.069 / 0.072 / 0.055&%
\scriptsize
0.069 / 0.072 / 0.054&%
\scriptsize
0.068 / 0.072 / 0.053&%
\scriptsize
\textbf{0.065} / \textbf{0.069} / \textbf{0.040}\\%
\rotatebox{90}{\begin{minipage}{6em}\centering SC-I\\(2434\\images)\end{minipage}}&%
\includegraphics[trim={12px 12px 12px 12px},clip,width=\imagewidth]{figures/model_evaluation_bias_free/Core_infrared_left/result_15px_central_opencv.jpg}&%
\includegraphics[trim={12px 12px 12px 12px},clip,width=\imagewidth]{figures/model_evaluation_bias_free/Core_infrared_left/result_15px_central_thin_prism_fisheye.jpg}&%
\includegraphics[trim={12px 12px 12px 12px},clip,width=\imagewidth]{figures/model_evaluation_bias_free/Core_infrared_left/result_15px_central_radial.jpg}&%
\figpic{Core_infrared_left/result_15px_central_generic_30.jpg}{2.5k}&%
\figpic{Core_infrared_left/result_15px_central_generic_20.jpg}{5.6k}&%
\figpic{Core_infrared_left/result_15px_central_generic_10.jpg}{21.8k}&%
\figpic{Core_infrared_left/result_15px_noncentral_generic_20.jpg}{14.0k}\\[-0.2em]%
Errors$^1$&%
\scriptsize
0.069 / 0.064 / 0.589&%
\scriptsize
0.053 / 0.046 / 0.440&%
\scriptsize
0.069 / 0.064 / 0.585&%
\scriptsize
0.035 / 0.030 / 0.133&%
\scriptsize
0.035 / 0.030 / 0.139&%
\scriptsize
0.034 / 0.030 / 0.137&%
\scriptsize
\textbf{0.030} / \textbf{0.026} / \textbf{0.120}\\%
\rotatebox{90}{\begin{minipage}{6em}\centering Tango\\(2293\\images)\end{minipage}}&%
\includegraphics[trim={12px 12px 12px 12px},clip,width=\imagewidth]{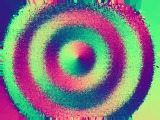}&%
\includegraphics[trim={12px 12px 12px 12px},clip,width=\imagewidth]{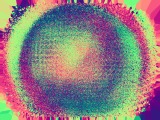}&%
\includegraphics[trim={12px 12px 12px 12px},clip,width=\imagewidth]{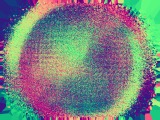}&%
\figpic{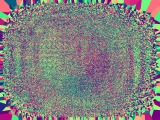}{0.7k}&%
\figpic{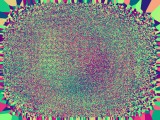}{1.6k}&%
\figpic{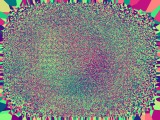}{6.0k}&%
\figpic{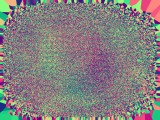}{4.0k}\\%
Errors$^1$&%
\scriptsize
0.067 / 0.062 / 0.776&%
\scriptsize
0.034 / 0.031 / 0.367&%
\scriptsize
0.033 / 0.029 / 0.331&%
\scriptsize
0.022 / 0.021 / 0.130&%
\scriptsize
0.022 / 0.021 / 0.127&%
\scriptsize
0.022 / 0.021 / \textbf{0.125}&%
\scriptsize
\textbf{0.020} / \textbf{0.019} / 0.131\\%
\rotatebox{90}{\begin{minipage}{5em}\centering FPGA\\(2142\\images)\end{minipage}}&%
\includegraphics[trim={12px 12px 12px 12px},clip,width=\imagewidth]{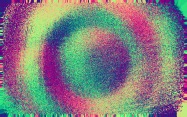}&%
\includegraphics[trim={12px 12px 12px 12px},clip,width=\imagewidth]{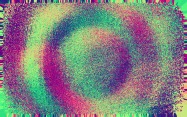}&%
\includegraphics[trim={12px 12px 12px 12px},clip,width=\imagewidth]{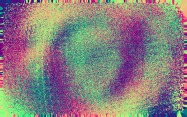}&%
\figpic{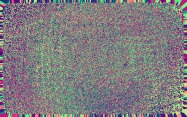}{0.8k}&%
\figpic{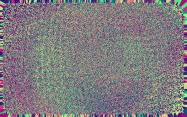}{1.8k}&%
\figpic{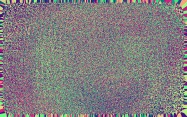}{6.8k}&%
\figpic{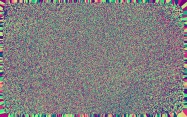}{4.6k}\\%
Errors$^1$&%
\scriptsize
0.024 / 0.022 / 0.442&%
\scriptsize
0.019 / 0.018 / 0.379&%
\scriptsize
0.021 / 0.019 / 0.317&%
\scriptsize
0.016 / 0.015 / 0.091&%
\scriptsize
0.016 / 0.015 / 0.091&%
\scriptsize
0.015 / 0.015 / 0.091&%
\scriptsize
\textbf{0.012} / \textbf{0.012} / \textbf{0.044}\\%
\\
&%
&%
&%
&%
Central Generic&%
Central Generic&%
Central Generic&%
Noncentral Generic\\%
&%
OpenCV&%
Thin-Prism Fisheye&%
Central Radial&%
ca.~60 px/cell&%
ca.~50 px/cell&%
ca.~40 px/cell&%
ca.~50 px/cell\\%
\rotatebox{90}{\begin{minipage}{6em}\centering GoPro\\(440\\images)\end{minipage}}&%
\includegraphics[trim={12px 12px 12px 12px},clip,width=\imagewidth]{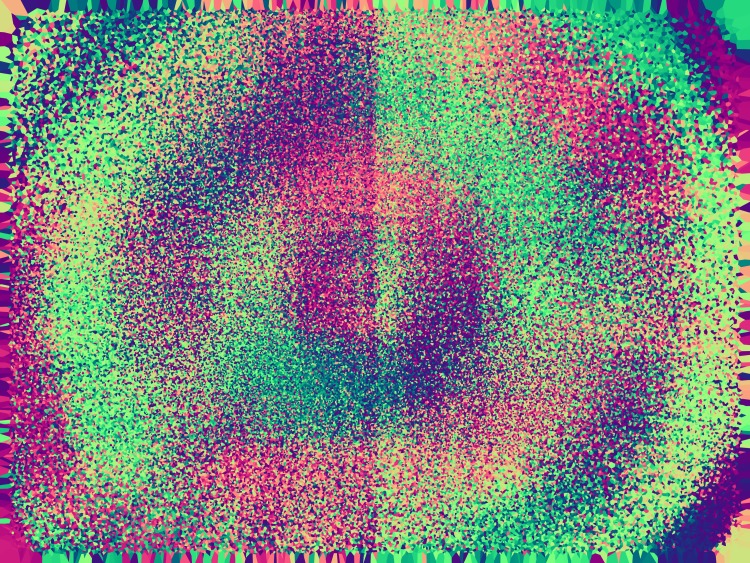}&%
\includegraphics[trim={12px 12px 12px 12px},clip,width=\imagewidth]{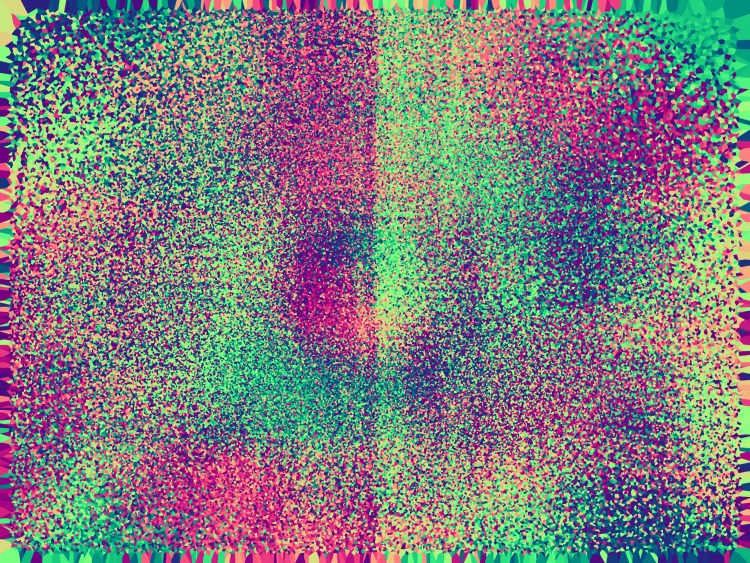}&%
\includegraphics[trim={12px 12px 12px 12px},clip,width=\imagewidth]{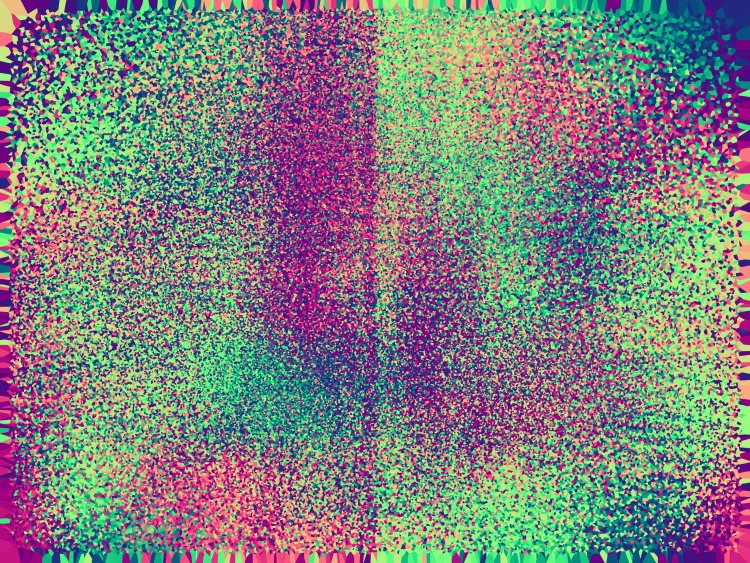}%
&%
\figpic{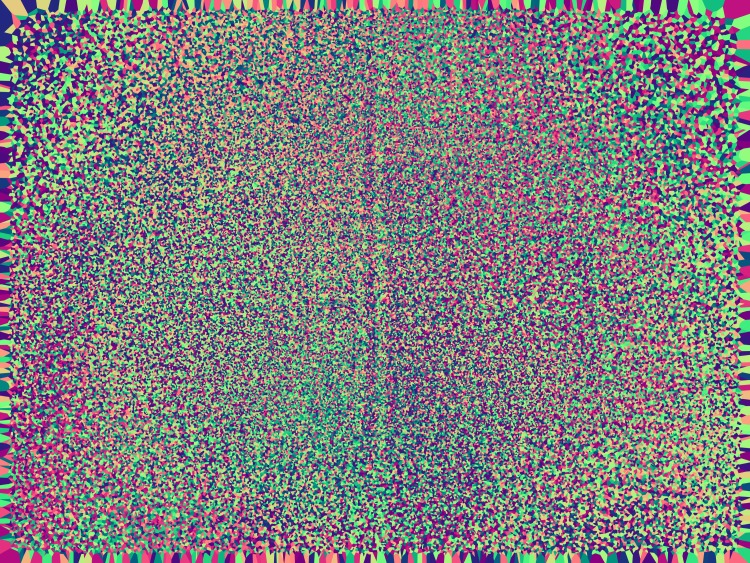}{3.9k}&%
\figpic{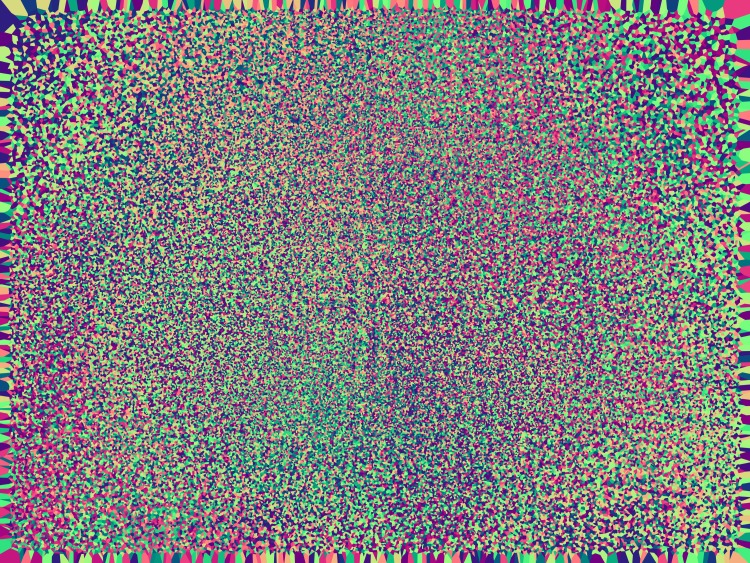}{5.4k}&%
\figpic{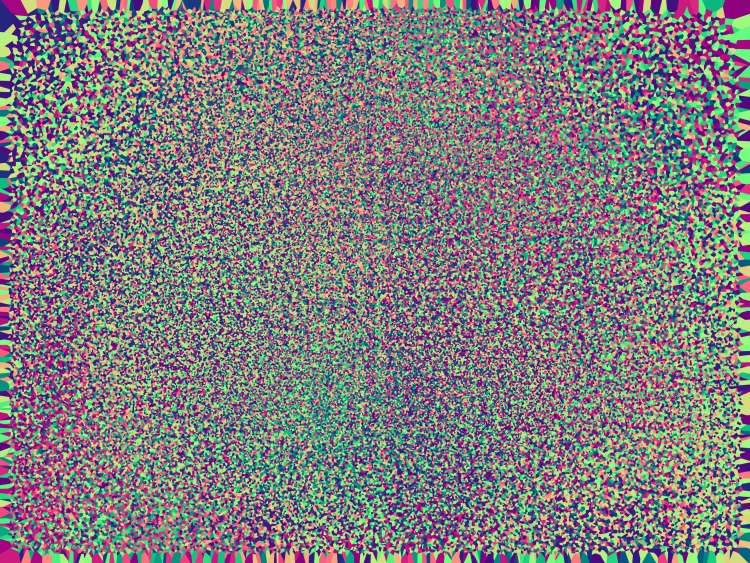}{8.4k}&%
\figpic{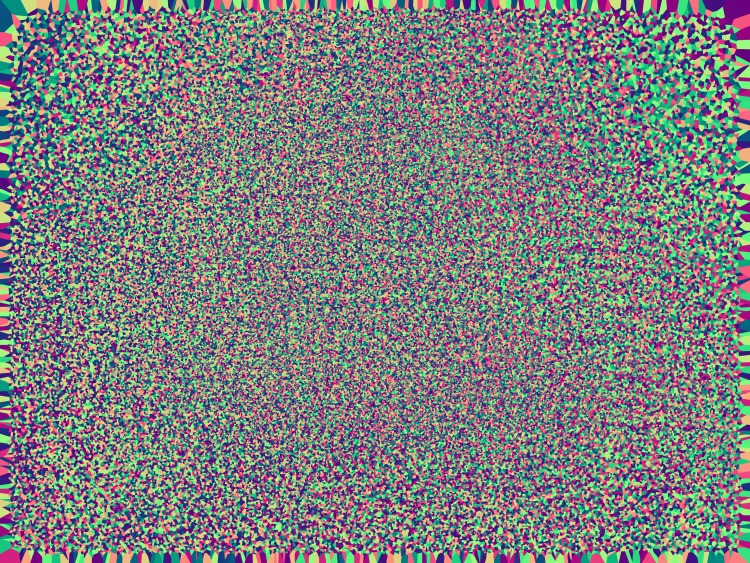}{13.5k}\\%
Errors$^1$&%
\scriptsize
0.113 / 0.115 / 0.819&%
\scriptsize
0.105 / 0.108 / 0.773&%
\scriptsize
0.106 / 0.108 / 0.759&%
\scriptsize
0.091 / 0.095 / 0.676&%
\scriptsize
0.091 / 0.095 / 0.679&%
\scriptsize
0.091 / 0.096 / 0.672&%
\scriptsize
\textbf{0.060} / \textbf{0.066} / \textbf{0.642}\\%
\rotatebox{90}{\begin{minipage}{6em}\centering HTC One M9\\(195\\images)\end{minipage}}&%
\includegraphics[trim={12px 12px 12px 12px},clip,width=\imagewidth]{figures/model_evaluation_bias_free/HTCOneM9/result_15px_central_opencv.jpg}&%
\includegraphics[trim={12px 12px 12px 12px},clip,width=\imagewidth]{figures/model_evaluation_bias_free/HTCOneM9/result_15px_central_thin_prism_fisheye.jpg}&%
\includegraphics[trim={12px 12px 12px 12px},clip,width=\imagewidth]{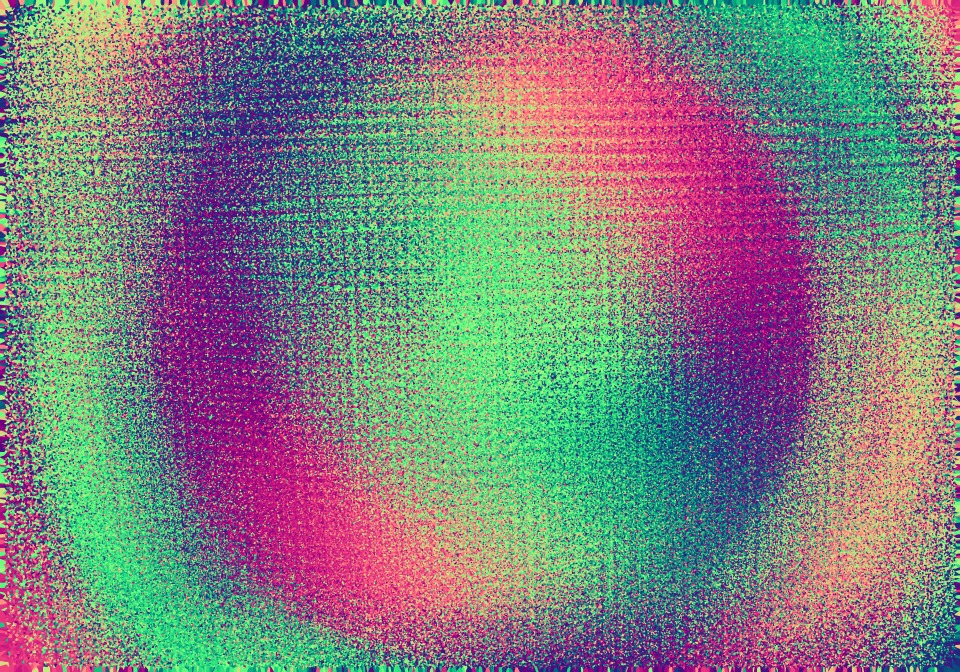}%
&%
\figpic{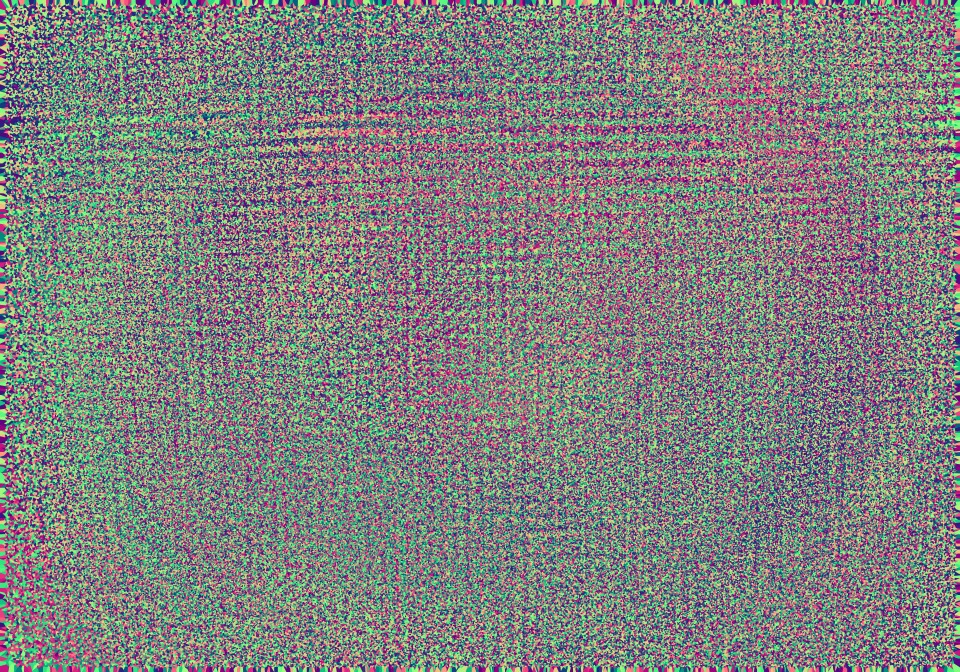}{6.0k}&%
\figpic{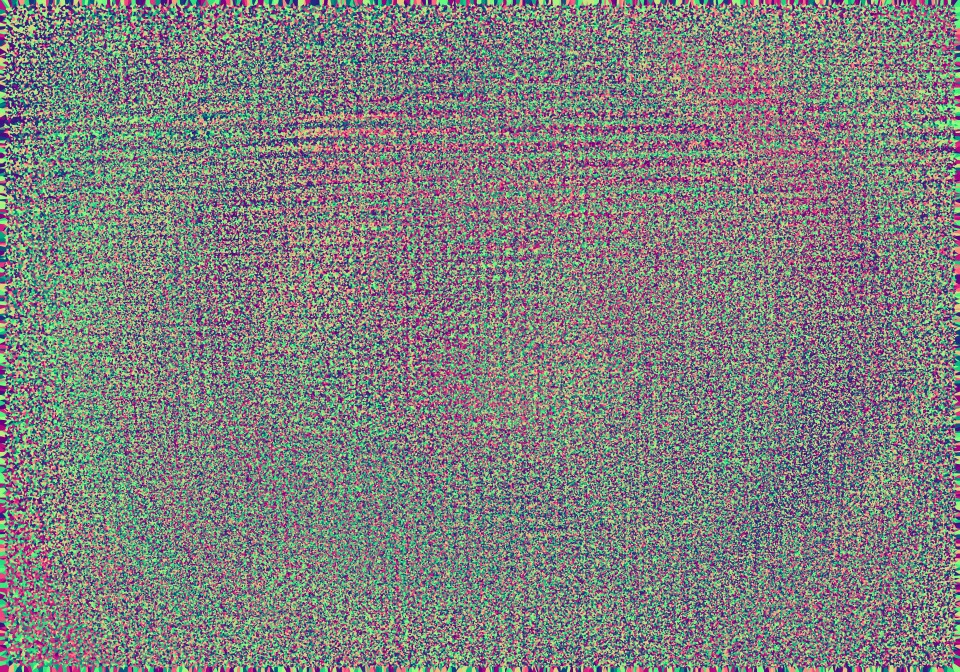}{8.6k}&%
\figpic{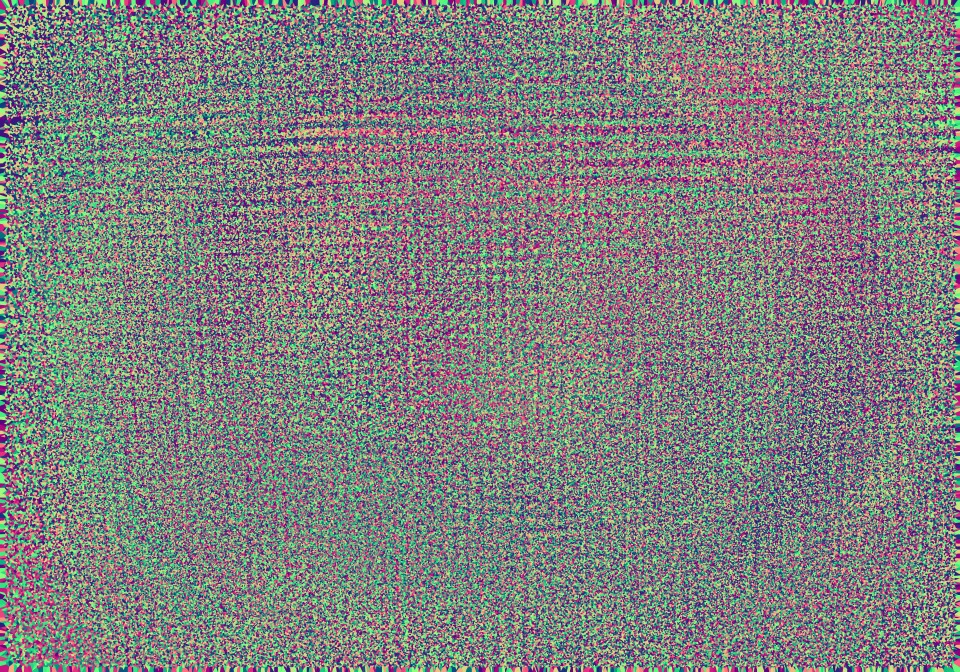}{13.2k}&%
\figpic{HTCOneM9/result_15px_noncentral_generic_50.jpg}{21.4k}\\%
Errors$^1$&%
\scriptsize
0.178 / 0.161 / 1.174&%
\scriptsize
0.360 / 0.298 / 1.830&%
\scriptsize
0.089 / 0.095 / 0.694&%
\scriptsize
0.043 / 0.045 / 0.378&%
\scriptsize
0.043 / 0.045 / 0.378&%
\scriptsize
0.043 / 0.045 / 0.377&%
\scriptsize
\textbf{0.039} / \textbf{0.039} / \textbf{0.352}\\%
\end{tabular}
\end{center}
\begin{minipage}[c]{0.14\linewidth}
\includegraphics[width=2.15cm]{figures/model_evaluation_bias_free/legend_error_directions.jpg}%
\vspace{-0.75em}
\end{minipage}%
\hfill%
\begin{minipage}[c]{0.86\linewidth}
\caption[]{
Directions (encoded as colors, see legend on the left) of all reprojection errors for calibrating the camera defined by the row with the model defined by the column.
Each pixel shows the direction of the closest reprojection error (\ie, the images are Voronoi diagrams) from all used images.
Ideally, the result is free from any systematic pattern.
Patterns indicate biased results arising from not being able to model the true camera.
Parameter counts for generic models are given in the images.
\hfill $^1$Median training error [px] / test error [px] / biasedness.
\vspace{-1em}
}\label{fig:model_eval_bias_free_larger_images}
\end{minipage}
\end{figure*}

Fig.~\ref{fig:model_eval_bias_free_larger_images} extends Fig.~\ref{fig:model_eval_bias_free} in the paper with results for additional cameras.
The specifications of all cameras in the figure are given in Tab.~\ref{tab:evaluation_cameras_supp}.
Note that the ``Tango'' camera has a fisheye lens and shows hardly any image information in the corners.
Thus, there are no feature detections in the image corners, which causes large Voronoi cells to be there in Fig.~\ref{fig:model_eval_bias_free_larger_images}.

\section{Example Application: Camera Pose Estimation}
\label{sec:application_example_localization_supp}
\begin{figure*}
\includegraphics[width=1.0\linewidth]{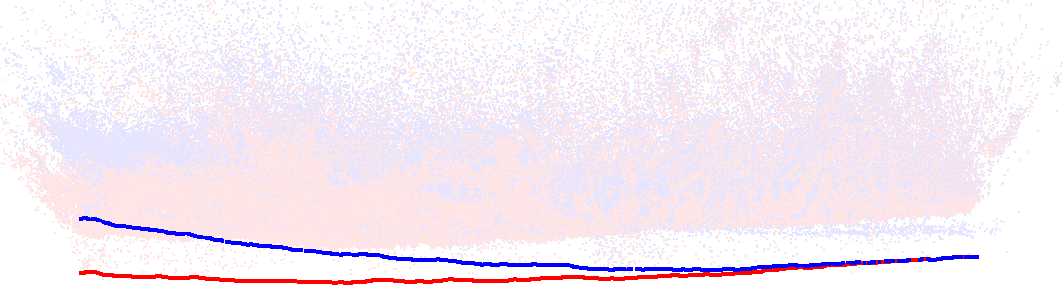}\\
\includegraphics[width=1.0\linewidth]{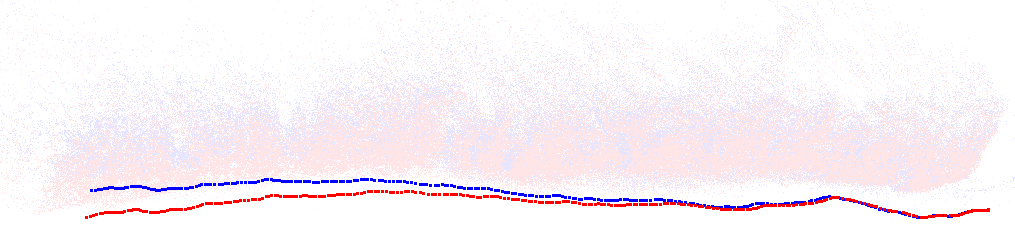}\\
\hrule
\includegraphics[width=1.0\linewidth]{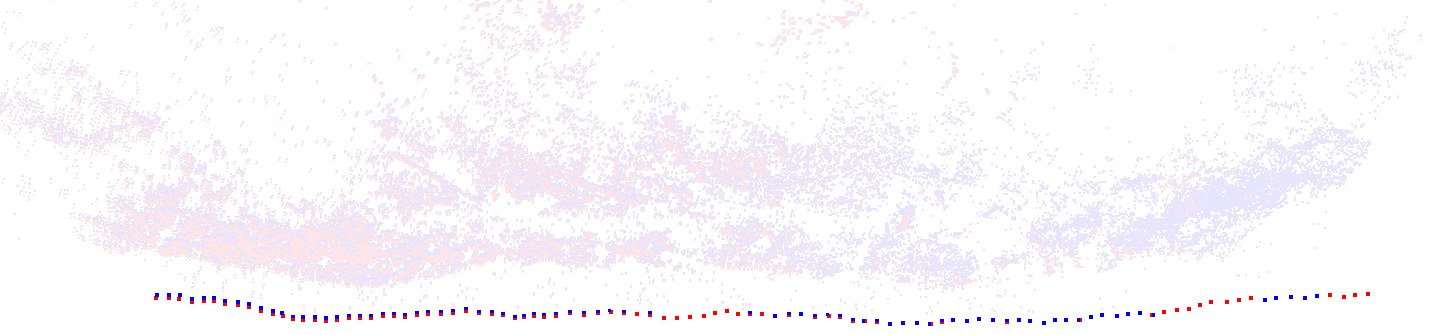}\\
\includegraphics[width=1.0\linewidth]{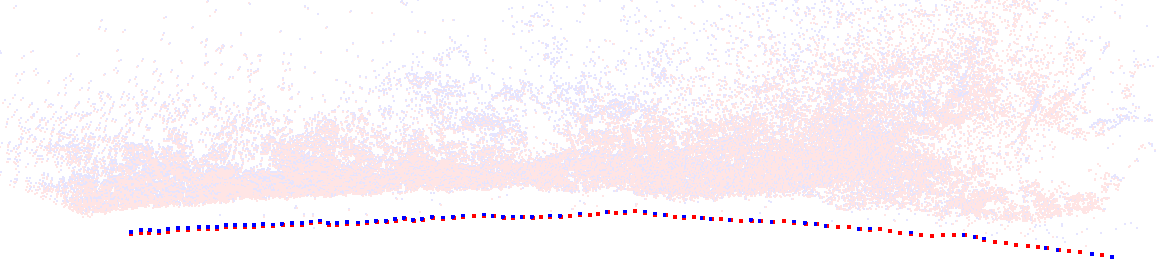}\\
\caption[]{
Example reconstructions of forest scenes used for the Structure-from-Motion experiment in the paper, bundle-adjusted with a noncentral-generic and a Thin-Prism-Fisheye calibration.
For the noncentral-generic model, camera poses are shown in blue and reconstructed points in light blue.
For the Thin-Prism-Fisheye model, camera poses are shown in red and reconstructed points in light red.
The two images on the top show reconstructions of videos by the SC-C camera, while
the two images on the bottom show reconstructions of videos by the D435-I camera.
The reconstructions are aligned at the first camera pose of the video (on the right side) to visualize the accumulated difference when starting from the same pose.
\vspace{-1em}
}\label{fig:structure_from_motion_reconstructions}
\end{figure*}

In Fig.~\ref{fig:structure_from_motion_reconstructions}, we present example images of the sparse 3D reconstructions that we evaluated in the paper to determine how much the results of bundle adjustment in Structure-from-Motion are affected by the choice of camera model.
The videos used for these reconstructions were recorded with the SC-C and D435-I cameras by walking in a mostly straight line on a forest road, looking either sideways or forward.
For D435-I, we use 10 videos with 100 frames each, whereas for SC-C, we use 7 videos with 264 frames on average.

\newpage
~
\newpage
~
\newpage
~
\newpage
~
\newpage

{\small
\bibliographystyle{ieee_fullname}
\bibliography{bib}
}

\end{document}